\documentclass[sigconf]{acmart_arxiv}
\usepackage{multirow}
\definecolor{cvprblue}{rgb}{0.21,0.49,0.74}
\definecolor{lightblue}{RGB}{57,113,181}
\usepackage{graphicx}
\usepackage{subcaption}
\usepackage[linesnumbered,ruled,lined]{algorithm2e}
\usepackage{array} 
\usepackage{amsfonts}
\usepackage{relsize}
\usepackage{xcolor}
\usepackage[numbers]{natbib}
\newcommand{\myhline}{\noalign{\global\arrayrulewidth0.3mm}\hline
                      \noalign{\global\arrayrulewidth0.3pt}}
\usepackage{makecell}

\newcommand{\method}{FedELMY}

\AtBeginDocument{%
  \providecommand\BibTeX{{%
    \normalfont B\kern-0.5em{\scshape i\kern-0.25em b}\kern-0.8em\TeX}}}

\setcopyright{none}
\makeatletter
\renewcommand\@formatdoi[1]{\ignorespaces}
\makeatother
\renewcommand{\footnotetextcopyrightpermission}[1]{}
\begin{document}

\title{One-Shot Sequential Federated Learning for Non-IID Data by Enhancing Local Model Diversity}


\author{Naibo Wang}
\affiliation{%
  \institution{National University of Singapore}
  \city{Singapore}
  \country{Singapore}}
\email{naibowang@comp.nus.edu.sg}

\author{Yuchen Deng}
\affiliation{%
  \institution{National University of Singapore}
  \city{Singapore}
  \country{Singapore}}
\email{dengyuchen.cc@gmail.com}

\author{Wenjie Feng}
\affiliation{%
  \institution{National University of Singapore}
  \city{Singapore}
  \country{Singapore}}
\email{wenjie.feng@nus.edu.sg}

\author{Shichen Fan}
\affiliation{%
  \institution{Xidian University}
  \city{Xi'an}
  \country{China}}
\email{shichenfan@stu.xidian.edu.cn}

\author{Jianwei Yin}
\affiliation{%
  \institution{Zhejiang University}
  \city{Hangzhou}
  \country{China}}
\email{zjuyjw@cs.zju.edu.cn}

\author{See-Kiong Ng}
\affiliation{%
  \institution{National University of Singapore}
  \city{Singapore}
  \country{Singapore}}
\email{seekiong@nus.edu.sg}
\renewcommand{\shortauthors}{author name and author name, et al.}

\begin{abstract}
  Traditional federated learning mainly focuses on parallel settings (PFL), which can suffer significant communication and computation costs. In contrast, one-shot and sequential federated learning (SFL) have emerged as innovative paradigms to alleviate these costs. However, the issue of non-IID (Independent and Identically Distributed) data persists as a significant challenge in one-shot and SFL settings, exacerbated by the restricted communication between clients. In this paper, we improve the one-shot sequential federated learning for non-IID data by proposing a local model diversity-enhancing strategy. Specifically, to leverage the potential of local model diversity for improving model performance, we introduce a local model pool for each client that comprises diverse models generated during local training, and propose two distance measurements to further enhance the model diversity and mitigate the effect of non-IID data. Consequently, our proposed framework can improve the global model performance while maintaining low communication costs. Extensive experiments demonstrate that our method exhibits superior performance to existing one-shot PFL methods and achieves better accuracy compared with state-of-the-art one-shot SFL methods on both label-skew and domain-shift tasks (e.g., 6\%+ accuracy improvement on the CIFAR-10 dataset).
\end{abstract}






\maketitle

\section{Introduction}
Federated learning (FL)~\cite{mcmahan2017communication} is a promising paradigm which enables collaborative machine
learning~\cite{wang2022collaborative, balkus2022survey} amongst multiple clients to build a consensus global model without the need to access others’ datasets. This paradigm offers salient benefits such as preservation of privacy~\cite{gao2021privacy}, security of data~\cite{ma2020safeguarding}, and the facility for different clients to derive a model exhibiting a higher degree of inference capability compared to individual client-based training~\cite{yang2019federated}. 

\begin{figure}[t]
    \centering
    \begin{tabular}{@{\extracolsep{\fill}}c@{}c@{\extracolsep{\fill}}}
            \includegraphics[width=0.5\linewidth]{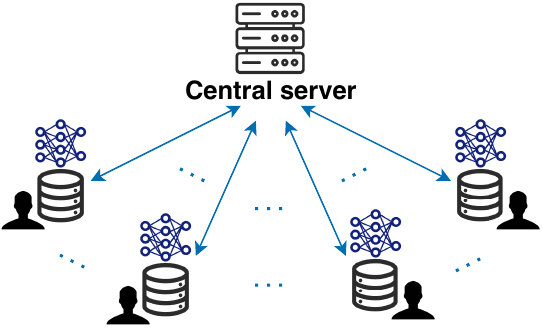} &
            \includegraphics[width=0.45\linewidth]{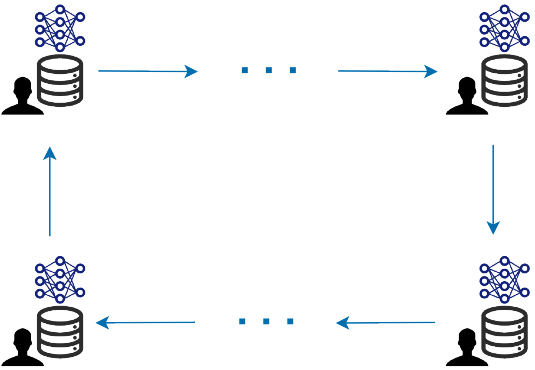}\\
            \begin{minipage}[t]{0.49\linewidth}
            \centering
            \textbf{(a) Parallel FL framework}
            \end{minipage} & 
            \begin{minipage}[t]{0.49\linewidth}
            \centering
            \textbf{(b) Sequential FL framework}
            \end{minipage}
    \end{tabular}
    \caption{Two federated learning settings.}
    \label{fig:intro}
\end{figure}

As shown in Fig. \ref{fig:intro}, two federated learning paradigms are parallel FL (PFL)~\cite{bonawitz2019towards, qu2022rethinking} and sequential FL (SFL)~\cite{li2024convergence}. PFL synchronizes model training across clients in parallel, such as FedAvg~\cite{mcmahan2017communication}, while SFL adopts a client-by-client training sequence that is widely applied in many scenarios such as healthcare~\cite{chen2023metafed, chang2018distributed}. Compared with PFL, SFL exhibits significant advantages in training efficiency, as evidenced by a reduction in training rounds \cite{zaccone2022speeding}. SFL also shows robust performance with limited datasets \cite{kamp2023federated} and offers enhanced data privacy protection due to its decentralized architecture \cite{hegedHus2021decentralized}.

One main concern of SFL is communication costs. To mitigate communication costs, one-shot federated learning~\cite{guha2018one} has been proposed where only one communication round is needed for the clients to interact with the server or other clients. However, current one-shot FL works~\cite{zhang2022dense, heinbaugh2023data} mainly focus on PFL (Fig. \ref{fig:intro} (a)) which typically requires a central server to facilitate model training. This can lead to numerous limitations including risk of privacy leakage~\cite{kairouz2021advances} and server node bottlenecks~\cite{hard2021jointly}. To address these issues, the decentralized structure~\cite{su2023one} has been incorporated into one-shot PFL which typically utilizes a mesh-topology network for clients to disseminate their models to others.
Nonetheless, this setting still leads to significant communication overhead than SFL.
In contrast, one-shot SFL, wherein each client only needs to communicate with its adjacent client once, can greatly reduce communication costs. However, existing studies on one-shot SFL~\cite{sheller2020federated, chang2018distributed} struggle with handling the non-IID data, a common challenge in FL that significantly impairs the performance of federated models~\cite{shoham2019overcoming}. How to tackle the non-IID data in one-shot SFL is still an open problem.

\begin{figure*}[t]
  \centering
  \begin{subfigure}{0.365\linewidth}
    \includegraphics[width=\linewidth]{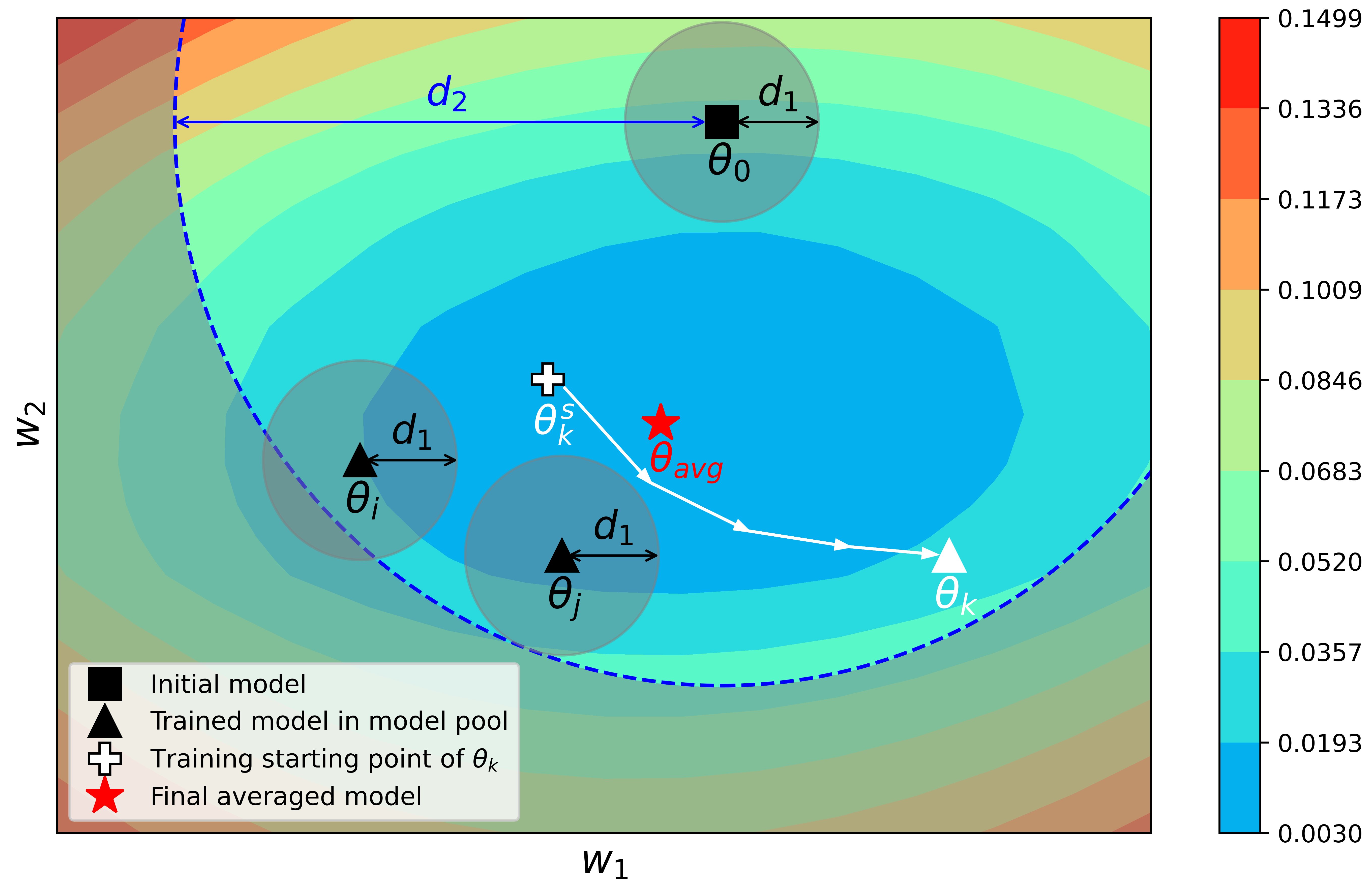}
    \caption{Training loss surface on local dataset $D_i$.}
  \end{subfigure}
  \begin{subfigure}{0.36\linewidth}
    \includegraphics[width=\linewidth]{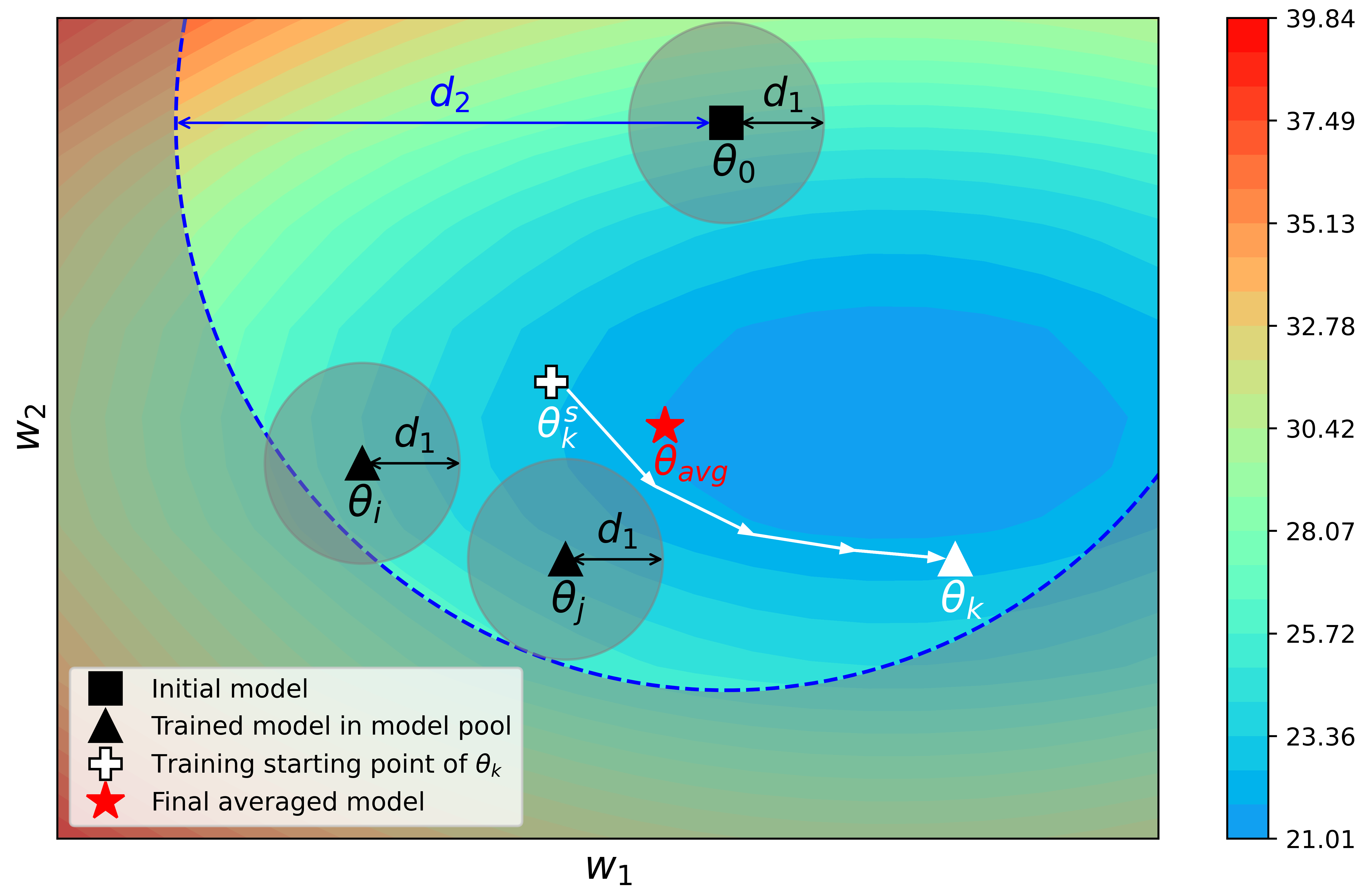}
    \caption{Test error surface on the whole test set.}
  \end{subfigure}
\caption{An illustration of our training solution on client $i$. Based on previously trained models in the model pool, every new model $m_k$ starts training from $\theta_k^s=f(\{\theta_i\}_{i=0}^{k-1})$ to improve training diversity ($f$ is the average function in our paper). During training, optimization of $\theta_k$ is constrained within a specific region (the non-shadow areas). $\theta_k$ is required to maintain a certain distance ($d_1$) from existing models $\{\theta_i\}_{i=0}^{k-1}$ to enhance model diversity, and should not diverge significantly ($d_2$) from the initial model $\theta_0$ to prevent deviation from the globally optimal solution caused by non-IID data. After training, all trained models $\{\theta_i\}_{i=1}^{k}$ display similar training losses on the local dataset $D_i$ (a) but have different test errors on the whole test set (b). Meanwhile, the averaged model $\theta_{avg}$ of all models in the model pool achieves a lower test error than any single model (b). }
   \label{fig:idea}
\end{figure*}

The key question in one-shot SFL is how to effectively transfer knowledge given the limited communication, especially with the non-IID data. Our insight is to utilize the diversity of models. Existing studies have shown that combining multiple networks can notably enhance the model's performance due to the inherent diversity amongst model weights~\cite{wortsman2022model, cha2021swad, izmailov2018averaging}. This observation has also been solidly backed by theoretical support from Rame et al.~\cite{rame2022diverse}. However, directly applying such a diversity strategy in SFL is not feasible, since SFL restricts each client to receive only one model from its adjacent client. This limitation could lead to insufficient diversity during model training, which can potentially undermine the performance of the final trained model. To overcome this challenge within the one-shot SFL framework, we propose a diversity-enhanced mechanism for model training. This mechanism is designed to augment the diversity of locally trained models. By generating a broader spectrum of models, we facilitate enriched knowledge transfer between adjacent clients, which also serves to mitigate the impact of data distribution disparities across clients.

In this paper, we present a novel one-shot sequential \textit{Fed}erated learning framework by \textit{E}nhancing \textit{L}ocal \textit{M}odel diversit\textit{Y} (\method). Specifically, we improve the local model training diversity by constructing a \textit{model pool} consisting of various models for each client. We introduce two distance regularization terms during the local training process to enhance model training diversity while mitigating the impact of non-IID data. Fig. \ref{fig:idea} presents a case study illustrating our core concept. Compared with conventional federated learning methods, our approach effectively minimizes communication costs, safeguards data privacy, and mitigates the detrimental effects of non-IID data, thereby enhancing overall model performance. 
Experiments show that our method can outperform existing one-shot SFL methods on both label-skew and domain-shift datasets.


The main contributions of this paper are as follows:  

\begin{itemize}
    \item We tackle the novel and practical problem of one-shot sequential federated learning. To the best of our knowledge, this is the first work to systematically investigate the one-shot communication setting in sequential federated learning.
    \item We introduce a novel framework \method\ to reduce the communication cost and improve the global model performance by enhancing model diversity during local training.
    \item We conduct extensive experiments on four datasets, considering both feature and label distribution shifts. Our method achieves superior performance compared with existing one-shot PFL and SFL methods.
\end{itemize}


\section{Related Work}


\subsection{Parallel Federated Learning}

One well-known PFL method is FedAvg~\cite{mcmahan2017communication}, whose performance is hindered due to the dispersed nature of the data (non-IID). Methods such as FedProx~\cite{li2020federatedprox}, FedDyn~\cite{acar2021federated}, Astraea~\cite{duan2020self}, pFedMe~\cite{t2020personalized}, and FedCurv~\cite{shoham2019overcoming} use a measure of global parameter stiffness to tackle data heterogeneity in federated learning. 
Additionally, approaches like FedDC~\cite{gao2022feddc} and SCAFFOLD~\cite{karimireddy2020scaffold} use a global gradient adjustment mechanism to manage local variations of the data. Other methods like FedGMM~\cite{wu2023personalized}, FCCL~\cite{huang2022learn}, FedBN~\cite{li2021fedbn} and ADCOL~\cite{li2023adversarial} utilize the personalized model instead of a shared global model to improve the performance of federated learning. These methods all necessitate a central server for model training. 



Another PFL scenario is the decentralized federated learning~\cite{yuan2023decentralized, beltran2023decentralized, roy2019braintorrent} when edge devices conduct training without a central server. 
Sun et al.~\cite{sun2022decentralized} proposed the decentralized FedAvg with momentum (DFedAvgM) algorithm to improve the performance of trained models in the decentralized federated learning setting. 
Shi et al.~\cite{shi2023improving} applied the SAM~\cite{foret2021sharpnessaware} optimizer to improve the model consistency of decentralized federated learning. This setting can better protect privacy than centralized PFL but incurs more communication costs.





\subsection{Sequential Federated Learning}

Recently, Sequential Federated Learning (SFL) started to gain attention in the FL community. Micah J et al.~\cite{sheller2020federated}
proposed a basic SFL framework to facilitate multi-institutional collaborations without sharing patient data; 
Li et al.~\cite{li2024convergence} proved that the convergence guarantees of SFL on heterogeneous data are better than PFL for both full and partial client participation, and validated that SFL outperforms PFL on extremely heterogeneous data in cross-device settings; 
Chen et al.~\cite{chen2023metafed} proposed MetaFed, an SFL scheme with cyclic knowledge distillation for personalized healthcare; 
Cho et al.~\cite{cho2023convergence} also provided convergence analysis of sequential federated averaging and proves that it can achieve a faster asymptotic convergence rate than vanilla FedAvg with uniform client participation under suitable conditions. 
However, existing works still lack the ability to deal with non-IID data and exhibit poor performance in real-world applications.



\subsection{One-shot Federated Learning}

One-shot federated learning~\cite{duan2023towards, guha2018one, su2023one, feng2023learning}, which aims to train a global model with only one round of communication between the clients and the server, has been proposed to reduce the communication cost and simultaneously enhance data privacy protection in federated learning. 
FedDISC \citep{yang2023exploring} made progress in one-shot semi-supervised federated learning, employing a pre-trained diffusion model. 
Investigations by Diao et al. \cite{diao2022towards} and Joshi et al. \cite{jhunjhunwala2023towards} have independently explored one-shot federated learning, viewing the problem through the respective perspectives of the open-set problem and Fisher information. Moreover, the proliferation of pre-trained models has sparked interest in collaborative, model-centric machine learning \cite{bommasani2021opportunities, wang2023data}. Instances such as FedKT~\cite{ijcai2021p205} and DENSE~\cite{zhang2022dense} are two one-shot federated learning schemes that choose to send models instead of gradients to the central server and generate a final global model through knowledge distillation. 
Despite their advancements, these one-shot methods are not designed for sequential federated learning, which still results in a higher risk of privacy leakage. Thus, devising a one-shot sequential federated learning framework is still under-explored.

\section{Method}

\begin{figure*}[t]
    \centering
     \includegraphics[width=\linewidth]{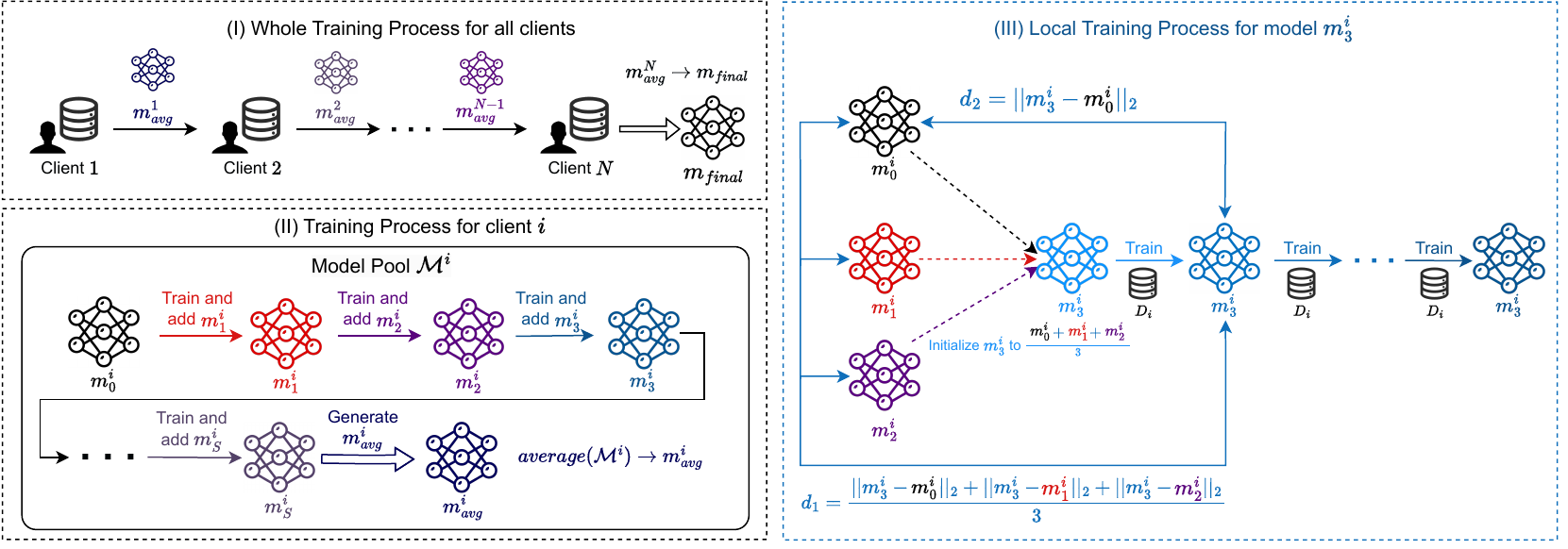}
     \caption{Overview of our method. Every client $i$ receives a model $m_{avg}^{i-1}$ from its previous client $i-1$ and sends model $m_{avg}^{i}$ to its next client $i+1$ after training (I). For each client $i$, we train $S$ models and put them into its model pool $\mathcal{M}^i$ (II). Every new model $m^i_j$ is initialized to the average of the existing models in $\mathcal{M}^i$ and trained under the control of $d_1$ and $d_2$  (III). }
     \label{fig:overview}
\end{figure*}

In this section, we introduce our problem setting and present our proposed algorithm \method. The overview of our method is demonstrated in Fig. \ref{fig:overview}.

\subsection{Problem Formulation}



Assume there are $N$ different clients (parties), each client has its own private dataset $D_i = \{(x_k, y_k)\}_{k=1}^{n_i}$ with size $n_i$. The principal goal of sequential federated learning is to develop a global model $m$ across the dataset $\mathcal{D} = \{D_i\}_{i=1}^{N}$, by minimizing the error on training data. The optimization objective is formulated as:

\begin{equation}
    \mathop{\min}\limits_{m} \sum\nolimits_{i=1}^{N} \mathbb{E}_{(x,y)\sim D_i}[L(m; x, y)],
\end{equation}


\noindent where $L(m; x, y)$ is the loss function evaluated on the private dataset $D_i$ from client $i$ ($c_i$) with model $m$. The training procedure of sequential federated learning is described as follows:

1. At the start of each training round $r$, the client's training sequence $\{c_{\pi_1}, c_{\pi_2}, \cdots, c_{\pi_N}\}$ is determined by randomly selecting indices $\{\pi_1, \pi_2, \cdots, \pi_N\}$ without replacement from set $\{1, 2, \cdots, N\}$.

2. At the very beginning of training when round $r=1$, randomly initialize global model $m^{(0)}$.

3. For the $i$-th client ($c_{\pi_i}$) in round $r$, initialize its model $m^{(r)}_{\pi_i,0}$ with the latest model:
\begin{equation}
m^{(r)}_{\pi_i,0} =
\begin{cases}
m^{(r-1)}, & \text{if } i=1 \\
m^{(r)}_{\pi_{i-1},E_{local}}, & \text{if } i>1
\end{cases}
\end{equation}

\noindent where $m^{(r-1)}$ is the global model received from round $r-1$, and $m^{(r)}_{\pi_{i-1}, E_{local}}$ is the model received from the $(i-1)$-th client ($c_{\pi_{i-1}}$) in round $r$ after it trained its model for $E_{local}$ epochs.

4. Update model $m^{(r)}_{\pi_i,0}$ for $E_{local}$ epochs based on $D_{\pi_i}$ and send $m^{(r)}_{\pi_i, E_{local}}$ to the $(i+1)$-th client ($c_{\pi_{i+1}}$), if SGD~\cite{bottou2010large} is chosen as the optimizer, then the update can be described by: 

\begin{equation}
    m^{(r)}_{\pi_i,k+1} = m^{(r)}_{\pi_i,k} - \eta \cdot g^{(r)}_{\pi_i,k} 
\end{equation}

\noindent where $m^{(r)}_{\pi_i,k+1}$ denotes the local model of client $\pi_i$ after $k$ local training steps in round $r$, $\eta$ is the learning rate, and $g^{(r)}_{\pi_i,k}$ represents the gradient of the loss function based on $D_{\pi_i}$ at step $k$.

5. At the end of round $r$, i.e., after all clients finished their training, we can get the global model $m^{(r)}$ of round $r$ as:

\begin{equation}
    m^{(r)} = m^{(r)}_{\pi_N, E_{local}}
\end{equation}

For one-shot sequential federated learning, only 1 training round is required, i.e., $r\equiv1$, thus $m^{(1)}$ will be the final global model.

\emph{Analysis.} Although the communication cost is low, the information exchanged in SFL is limited, especially under the one-shot setting. Therefore, it is a big challenge to aggregate diverse and useful information to get a better global model. In other words, the diversity of distributed data under the one-shot SFL paradigm has not been well utilized. Meanwhile, the non-IID data in SFL increases the risk of models getting stuck in a local minimum, which further limits the enhancement of the generalization ability of the global model~\cite{li2020federatedprox}. Hence, it is essential to explore methods to increase the diversity during model training and mitigate the impact of non-IID data to enhance the model's performance in one-shot SFL. 





\subsection{Local Model Diversity Enhancement}

\label{sec:model_pool}

The authors of SWAD~\cite{cha2021swad} demonstrate that solely minimizing empirical loss in a single model is typically insufficient for achieving good generalization. Furthermore, they argue that the performance markedly improves through averaging diverse models trained by various hyperparameters, even when the training losses of these models are similar. Therefore, we propose to enhance diversity by a \textit{M}odel \textit{P}ool, which is essentially a collection of models maintained by a client during its local training. Here, every client $i$ possesses a model pool $\mathcal{M}^i$ initially consisting of a solitary model, denoted as $m_0^i$. For the first client when $i=1$, the model $m_0^1$ undergoes the process of random initialization, followed by a \textit{warm-up} process over $E_{\mathrm{w}}$ epochs on its local dataset $D_1$. As for the subsequent clients indexed from $i=2:N$, each model, $m_0^i$, is assigned to be the average model $m_{avg}^{i-1}$ which is efficiently derived from the model pool $\mathcal{M}^{i-1}$ of the previous client $i-1$:


\begin{equation}
    m_{avg}^{i-1} \leftarrow\frac{\sum_{t=0}^{|\mathcal{M}^{i-1}|-1}m_t^{i-1}}{|\mathcal{M}^{i-1}|},
\end{equation}


The parameters to be sent to the next client have been improved to an average of a local diverse model pool instead of a single model.

Then, for each individual client $i$, we train an additional series of $S$ models, denoted as $\{m_j^i\}_{j=1}^{S}$, on the local dataset $D_i$ to probe a broader range of diverse models. Each new model, $m_j^i$, stems from the existing models in the established model pool, $\mathcal{M}^i$. More specifically, initialization of every new model $m_j^i$ is achieved by averaging all the models currently in circulation within the model pool $\mathcal{M}^i$, with the first model $m_0^i$ included; that is,
\vspace{-0.1cm}
\begin{equation}
m_j^i \leftarrow \frac{1}{|\mathcal{M}^i|}\sum\nolimits_{t=0}^{|\mathcal{M}^i|-1} m_t^i,
\end{equation}

 In this way, we craft a unique launching point for the training of each new model, ensuring a departure from all earlier models within our pool $\mathcal{M}^i$. Consequently, a diverse array of starting points is harnessed, granting us access to a more comprehensive set of potential solutions that evade the restrictions of a singular training trajectory and its attendant limitations.



After initialization, the model $m_j^i$ is then trained for $E_{local}$ epochs, employing the same hyperparameters as with preceding models. Upon finishing training, $m_j^i$ will be added to the model pool $\mathcal{M}^i$. 
After training all $S$ models for client $i$, the client will calculate the final averaged model $m_{avg}^{i}$, which is the average of all models in $\mathcal{M}^i$, and send $m_{avg}^{i}$ to the subsequent client, $i+1$. This process will be repeated until client $i=N$ concludes its local training. Upon completion, the final client $i=N$ will output the final model $m_{final}$, which is the average of all models in its model pool $\mathcal{M}^N$, i.e., $m_{final} \leftarrow \frac{1}{|\mathcal{M}^N|}\sum_{t=0}^{|\mathcal{M}^N|-1}m_t^N$.

To further enhance training diversity and prevent the convergence of the newly generated model $m_j^i$ towards the states of existing models in $\mathcal{M}^i$—thereby diminishing diversity—we aim to maintain the current exploration direction of the model as distinct as possible from those of existing models, throughout the training process. Consequently, we introduce a new distance control term, $d_1$, in the training process, which is defined as:
\begin{equation}
\label{eq:d1}
    \begin{split}
    d_1 &= \frac{1}{|\mathcal{M}^i|}\sum\nolimits_{t=0}^{|\mathcal{M}^i|-1} dist(m_j^i,m_t^i) \\
    &= \frac{1}{|\mathcal{M}^i|}\sum\nolimits_{t=0}^{|\mathcal{M}^i|-1} ||m_j^i-m_t^i||_2,
    \end{split}
\end{equation}
\noindent where $m_j^i$ represents the current model being trained for client $i$, and $m_t^i$ signifies a model in the model pool $\mathcal{M}^i$. This distance is calculated by averaging the $L_2-norm$ between the current model and all existing models in the model pool. The loss function incorporates and deducts this distance during training, allowing the model to maximize its distance from the existing models, therefore promoting training diversity.


\subsection{Mitigation of Non-IID Data Impact}

\label{sec:diversity_control}

In non-IID scenarios, variations in local objectives may risk deviating from the globally optimal solution across multiple local iterations, thereby impeding convergence~\cite{li2020federatedprox}. To prevent significant deviation of the model from the global solution, we further refine the model training process by introducing an additional regularization term $d_2$, defined as follows:
\begin{equation}
\label{eq:d2}
    d_2 = dist(m_j^i,m_0^i)=||m_j^i-m_0^i||_2,
\end{equation}



\noindent where $m_0^i$ represents the first model in the model pool $\mathcal{M}^i$ for client $i$. The loss function also incorporates this term to ensure that the model maintains reasonable proximity to the initial model $m_0^i$ (global solution of previous clients) in the pool during local updates. It helps mitigate the effect of non-IID data distribution while accommodating system heterogeneity.

The total loss function $\mathcal{L}$ for model $m_j^i$ is formed as follows:
\begin{equation}
\mathcal{L}(m_j^i)=\ell(m_j^i;D_i) - \alpha \cdot d_1 + \beta \cdot d_2,
\end{equation}

\noindent where $\ell$ denotes the original loss function, $D_i$ is the local dataset for client $i$, and $\alpha$ and $\beta$ are two hyperparameters that govern the effect of both distances on model training. Our aim is to achieve a balance that enhances model training diversity while mitigating the impact of non-IID data. 


\begin{algorithm}[t]
    \caption{Our proposed \method}
    \label{alg:method}
    \KwIn{Local datasets $\mathcal{D} = \{D_i\}_{i=1}^{N}$, warm-up epoch $E_{\mathrm{w}}$, learning rate $\eta$, number of local iterations $E_{local}$, model number to be trained per client $S$, scale hyperparameters $\alpha$, $\beta$}
    \KwOut{The final model $m_{final}$}
    
    \textbf{Initialization:} For client 1, warm up a randomly initialized model $m_{avg}^0$ for $E_{\mathrm{w}}$ epochs



        \For{client $i= 1:N$ }{
        
        Only for client $i= 2:N$, receive $m_{avg}^{i-1}$ from client $i-1$

        \textcolor{lightblue}{// Initialize model pool $\mathcal{M}^i$ for client $i$}

        $\mathcal{M}^i = \{m_0^i\}$ with $m_0^i \leftarrow$ $m_{avg}^{i-1}$
        
        \For{$j=1:S$}{
        
        \textcolor{lightblue}{// Initialize $m_j^i$}
        
        $m_j^i \leftarrow \frac{1}{|\mathcal{M}^i|}\sum_{t=0}^{|\mathcal{M}^i|-1} m_t^i$ 

        \textcolor{lightblue}{// Local training for $m_j^i$}
 
        \For{$k=1:E_{local}$}{

        
        


            
            
            $\mathcal{L}(m_j^i)=\ell(m_j^i;D_i) - \alpha \cdot d_1 + \beta \cdot d_2$

            $m_j^i \leftarrow m_j^i - \eta \nabla_m \mathcal{L}(m_j^i)$
        }

        $\mathcal{M}^i \leftarrow \mathcal{M}^i \cup \{m_j^i\}$
        
        }
        $m_{avg}^i \leftarrow\frac{1}{|\mathcal{M}^i|}\sum_{t=0}^{|\mathcal{M}^i|-1}m_t^i$  
        
        Only for client $i=1:N-1$, send $m_{avg}^i$ to client $i+1$

        }
        
        \textcolor{lightblue}{// For the final client $i=N$, output model $m_{final}$}
        
        $m_{final} \leftarrow m_{avg}^N$
\end{algorithm}
\vspace{-2mm}
\subsection{Implementation Details}
\label{sec:imp}

The details of our method are depicted in Algorithm \ref{alg:method}. The training procedure begins at client 1, which acts as the starting point for training with a warm-up phase (line 1).  Then, each client $i$ will create and initialize a model pool $\mathcal{M}^i$ (lines 3-5). The process of involving a new model $m_j^i$ in $\mathcal{M}^i$ is presented in line 8. Lines 10-13 elaborate on the local training process per model based on our customized loss function $\mathcal{L}$, as delineated in Sec. \ref{sec:diversity_control}. Once all models for client $i$ have completed training, the average model $m_{avg}^i$ derived from $\mathcal{M}^i$ is sent to the next client $i+1$ (lines 16-17). The final global model $m_{final}$ is obtained as the average of the models in model pool $\mathcal{M}^N$ of the final client $N$ (line 20).


\noindent\textbf{Communication cost.} In our method, each client will send a model to its adjacent client only once, thus only $N-1$ model exchanges are required. Therefore, the overall communication cost for all $N$ parties is $O(NM)$, where $M$ represents the size of the model.

\noindent \textbf{Computation cost.} In our approach, every client will train $S$ models, thus the overall computation cost for all $N$ parties is $O(NSE_{local})$. 

\section{Experiments}

\subsection{Experimental Setup}

\begin{figure}[t]
  \centering
  \begin{subfigure}{0.49\linewidth}
    \includegraphics[width=\linewidth]{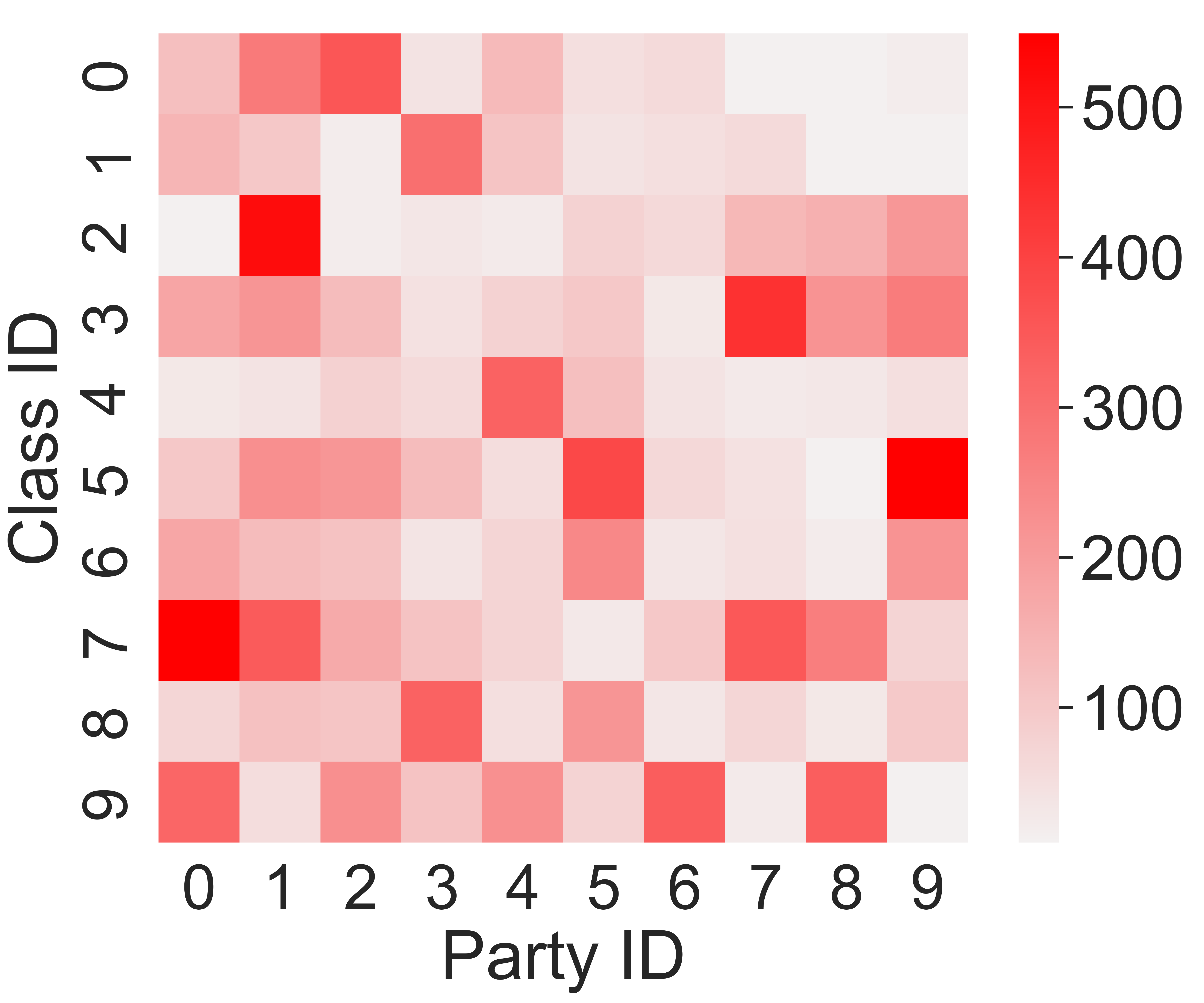}
    \caption{Label-skew distribution.}
  \end{subfigure}
  \hspace{0.5mm}
  \begin{subfigure}{0.46\linewidth}
    \includegraphics[width=\linewidth]{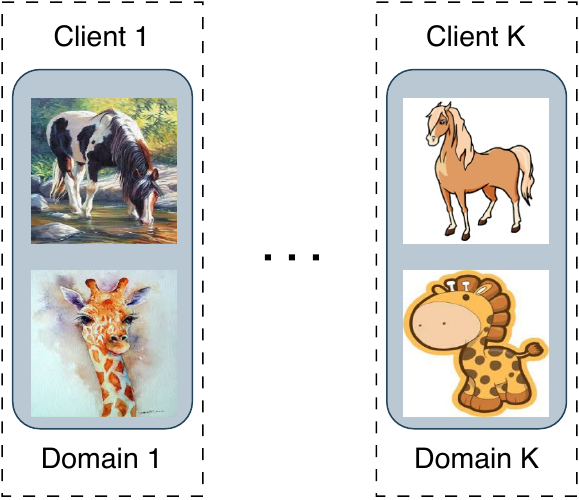}
    \caption{Domain-shift distribution.}
  \end{subfigure}
\caption{Two data distributions across clients. For the label-skew distribution, the color depth of every square represents the number of samples of the corresponding class on that client; for the domain-shift (feature-skew) distribution, every client possesses a specific domain with all classes.} 
   \label{fig:skew}
\end{figure}

\noindent\textbf{Datasets and Data Partition.} As shown in Fig. \ref{fig:skew}, in federated learning applications, non-IID data distribution is a common scenario, often arising in two forms: \textit{label-skew}~\cite{diao2022towards} and \textit{feature-skew} (also known as \textit{domain-shift})~\cite{huang2023rethinking}. We seek to corroborate the effectiveness of our method under these two distinctive non-IID conditions. We select two datasets associated with label-skew setting, namely \textbf{CIFAR-10}~\cite{krizhevsky2009learning} and \textbf{Tiny-ImageNet}~\cite{le2015tiny}, along with two datasets linked to domain-shift setting, specifically \textbf{PACS}~\cite{li2017deeper} and \textbf{Office-Caltech-10}~\cite{gong2012geodesic}. 

The training datasets for CIFAR-10 and Tiny-ImageNet are partitioned into $N=10$ clients with Dirichlet distribution $\beta=0.5$. 
Meanwhile, with respect to the PACS and Office-Caltech-10 datasets, they intrinsically contain four distinct domain samples, each allocated to a single client, yielding $N=4$ clients in total. 
Each client is provided with a randomly selected 90\% of the local training dataset for training purposes, with the remaining serving as the validation set. We test the final model $m_{final}$ on all test data from all clients, providing a global test performance measurement.

\begin{table*}[t]
    \centering
    \caption{Test accuracy (\%, mean$\pm$std) comparison of our \method\ method to other baselines on both label-skew and domain-shift (feature-skew) tasks. MetaFed, FedSeq and our \method\ are SFL methods, while other baselines are all PFL methods.}
\begin{tabular}{c|cc|cc|cc|cc}
\myhline
Distribution & \multicolumn{4}{c|}{Label-Skew} & \multicolumn{4}{c}{Domain-Shift} \\
\hline
Dataset & \multicolumn{2}{c|}{CIFAR-10} & \multicolumn{2}{c|}{Tiny-Imagenet} & \multicolumn{2}{c|}{PACS} & \multicolumn{2}{c}{Office-Caltech-10} \\
\hline
$E_{local}$ & 100   & 200   & 100   & 200   & 100   & 200   & 100   & 200 \\
\hline
DFedAvgM &   19.32$\pm$2.13    & 18.59$\pm$1.65 &   2.72$\pm$0.52    & 2.02$\pm$0.20 &  20.49$\pm$1.07     & 21.58$\pm$2.28 &    10.18$\pm$1.56   & 10.01$\pm$0.70 \\
DFedSAM &    17.12$\pm$0.44   & 18.51$\pm$1.28 &    2.58$\pm$0.29   & 3.15$\pm$0.29 &    19.01$\pm$1.09   & 20.79$\pm$1.36 &    14.74$\pm$1.63   & 15.09$\pm$1.04 \\
FedOV &   36.32$\pm$7.56    & 38.06$\pm$7.40 &     1.18$\pm$0.10 & 1.29$\pm$0.30 &    12.09$\pm$4.00  & 22.15$\pm$1.40 &   9.67$\pm$4.09    & 10.01$\pm$3.72 \\
DENSE & 61.76$\pm$0.43 & 64.59$\pm$1.04 & 1.27$\pm$0.45 & 1.49$\pm$0.05 & 17.34$\pm$2.14 & 15.60$\pm$5.40 & 33.12$\pm$5.16 & 36.68$\pm$4.87 \\
\hline
MetaFed & 71.46$\pm$0.95   &  71.31$\pm$0.73  &   23.52$\pm$0.29     &  24.76$\pm$0.24  &     35.67$\pm$0.27   &   38.73$\pm$2.61  &   41.78$\pm$3.24     &  42.26$\pm$3.77  \\
FedSeq & 72.92$\pm$0.62 & 73.59$\pm$0.95 & 25.50$\pm$0.34    & 25.08$\pm$0.89 &  43.68$\pm$1.29  & 46.53$\pm$0.75  &  32.78$\pm$3.99 & 37.07$\pm$5.78 \\
\method  &    \boldmath{}\textbf{79.03$\pm$0.74}\unboldmath{}    & \boldmath{}\textbf{80.28$\pm$1.18}\unboldmath{} &    \boldmath{}\textbf{32.84$\pm$0.22}\unboldmath{}   & \boldmath{}\textbf{30.42$\pm$0.10}\unboldmath{} &   \boldmath{}\textbf{46.08$\pm$1.82}\boldmath{}    & \boldmath{}\textbf{47.74$\pm$1.68}\unboldmath{} &    \boldmath{}\textbf{44.64$\pm$1.98}\boldmath{}   & \boldmath{}\textbf{45.26$\pm$3.93}\unboldmath{} \\
\myhline
\end{tabular}%

    \label{tab:main_results}
\end{table*}%

\noindent\textbf{Baselines.}
We compare our method with baselines under various federated learning settings: \textbf{DFedAvgM}~\cite{sun2022decentralized}, a decentralized adaptation of the extensively studied FedAvg~\cite{mcmahan2017communication} method in federated learning; \textbf{DFedSAM}~\cite{shi2023improving}, an application of the SAM~\cite{foret2021sharpnessaware} optimizer in decentralized PFL; \textbf{FedOV}~\cite{li2022learning}, a centralized one-shot federated learning method employing the open-set to bolster the final model; \textbf{DENSE}~\cite{zhang2022dense}, a one-shot federated learning framework also reliant on a central server to deliver a global model via knowledge distillation. Lastly, we compare our method to two sequential federated learning methods: \textbf{MetaFed}~\cite{chen2023metafed}, a personalized sequential federated learning approach used for healthcare, which requires at least two rounds of communication; \textbf{FedSeq}~\cite{li2024convergence}, which is the state-of-the-art sequential federated learning algorithm.



In an effort to ensure a fair comparison, we adapted the decentralized PFL methods DFedAvgM and DFedSAM to the one-shot setting, and adjusted these methods to select all clients for training and communication to fit the setting. In the remaining parts, unless otherwise specified, we set the local training epoch $E_{local}$ to 200. For more implementation details about our method and all baseline methods, such as the optimizer, learning rate and batch size we employed, please refer to Sec. 1 of the supplementary material.

\subsection{Effectiveness}
\label{sec:main_results}


\subsubsection{One-shot Setting}
Table~\ref{tab:main_results} presents the test accuracy of different methods in two training settings ($E_{local}=100$ and $200$) to verify the robustness of our method. Results show that our \method\ method outperforms all other methods across all datasets including both label-skew and domain-shift tasks. Specifically, \method\ surpasses existing SFL methods by over 6\% in accuracy, and outperforms current PFL methods by more than 15\% on the CIFAR-10 (label-skew) dataset. Additionally, \method\ achieves a 25\% higher accuracy over PFL methods and at least an 1.5\% improvement over SFL methods on the PACS (domain-shift) dataset. These results consistently demonstrate the effectiveness of our method.


It is noteworthy that for the Tiny-ImageNet dataset, due to its large number of classes (200), most PFL baselines fail to deliver effective performance such as DENSE (close to random guesses), suggesting that they cannot effectively handle datasets with too many classes. However, our method works well and also achieves improvement compared to other SFL methods. Such improvements are attributed to the diverse local training so that our method can learn better feature representations. 

In addition, it is not always the case that the test accuracy of $E_{local}=200$ is better than $100$, indicating that simply increasing the number of training rounds does not always improve model performance – enhancing diversity is an essential factor.

\begin{table}[t]
  \centering
  \renewcommand{\arraystretch}{1.1} 
  \caption{Few-Shot performance comparison for PACS dataset.}
    \begin{tabular}{ccccc}
    \myhline
    \textbf{Shot} & \textbf{1} & \textbf{3} &  \textbf{5} & \textbf{7} \\
    \hline
    MetaFed & 41.62\%  & 44.75\% &  46.11\%  &  46.64\% \\
    FedSeq & 47.31\% & 49.30\% & 50.96\% & 50.97\% \\
    \method  & \textbf{49.14\%} & \textbf{56.35\%} & \textbf{57.05\%} & \textbf{57.13\%} \\
    \myhline
    \end{tabular}%
  \label{tab:few_shot}%
\end{table}%

\subsubsection{Few-Shot Setting}

Although \method\ is designed for the one-shot setting, here we explore the performance of \method\ in few-shot scenarios to validate the scalability of our framework, i.e., when the final client $N$ sends model $m_{avg}^N$ to the first client $1$, thereby starting a new cycle of model training. As we can see from Table \ref{tab:few_shot}, even under few-shot settings, \method\ consistently outperforms the SFL baselines. Such observations underscore the effectiveness and scalability of \method. Meanwhile, we observe that as the number of training rounds increases to a certain extent, the overall performance does not significantly improve, indicating that the model has reached a state of convergence. This concludes that blindly increasing the number of training rounds will not significantly enhance the model performance.

\begin{figure}[t]
  \centering
   \includegraphics[width=.8\linewidth]{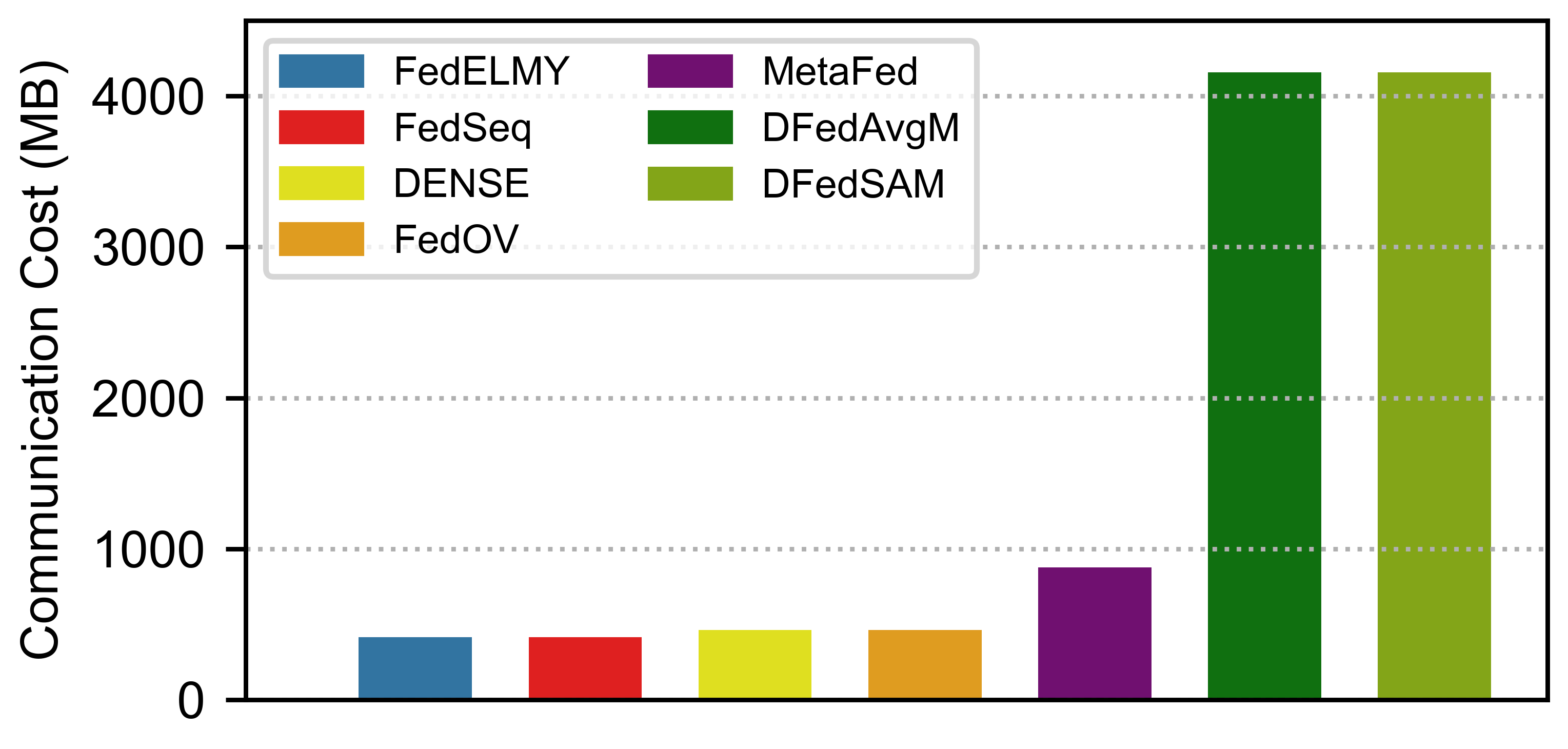}
   \caption{Communication cost comparison of different algorithms for CIFAR-10 dataset when the number of clients $N=10$ with model ResNet-18.}
   \label{fig:communication_cost}
\end{figure}

\subsection{Efficiency}


In this section, we will evaluate our method's efficiency by comparing its communication cost and model performance under different computation costs with other baselines.

\subsubsection{Communication cost.} 

Fig. \ref{fig:communication_cost} shows the communication expenses of different methods with $N = 10$ clients on the CIFAR-10 dataset with the Resnet-18 model, whose size is $M =$ 46.2 MB. The communication cost of \method\ is restricted to $(N-1) \times M =$ 415.8 MB, with only the FedSeq method displaying the same cost as ours; however, it notably underperforms in performance (Table. \ref{tab:main_results}). MetaFed requires $(2N-1) \times M =$ 877.8 MB of communication cost since it requires at least two rounds of communication (common knowledge accumulation and personalization). Central server-dependent methods, such as DENSE and FedOV, require an expenditure of $N \times M =$ 462 MB to transfer models to the server. Other methods mandating communication with their neighbors like DFedAvgM and DFedSAM, will require even higher communication costs. Accordingly, our method effectively diminishes the communication burden of FL and thus strengthens data privacy.

\begin{figure}[t]
  \centering
   \includegraphics[width=.95\linewidth]{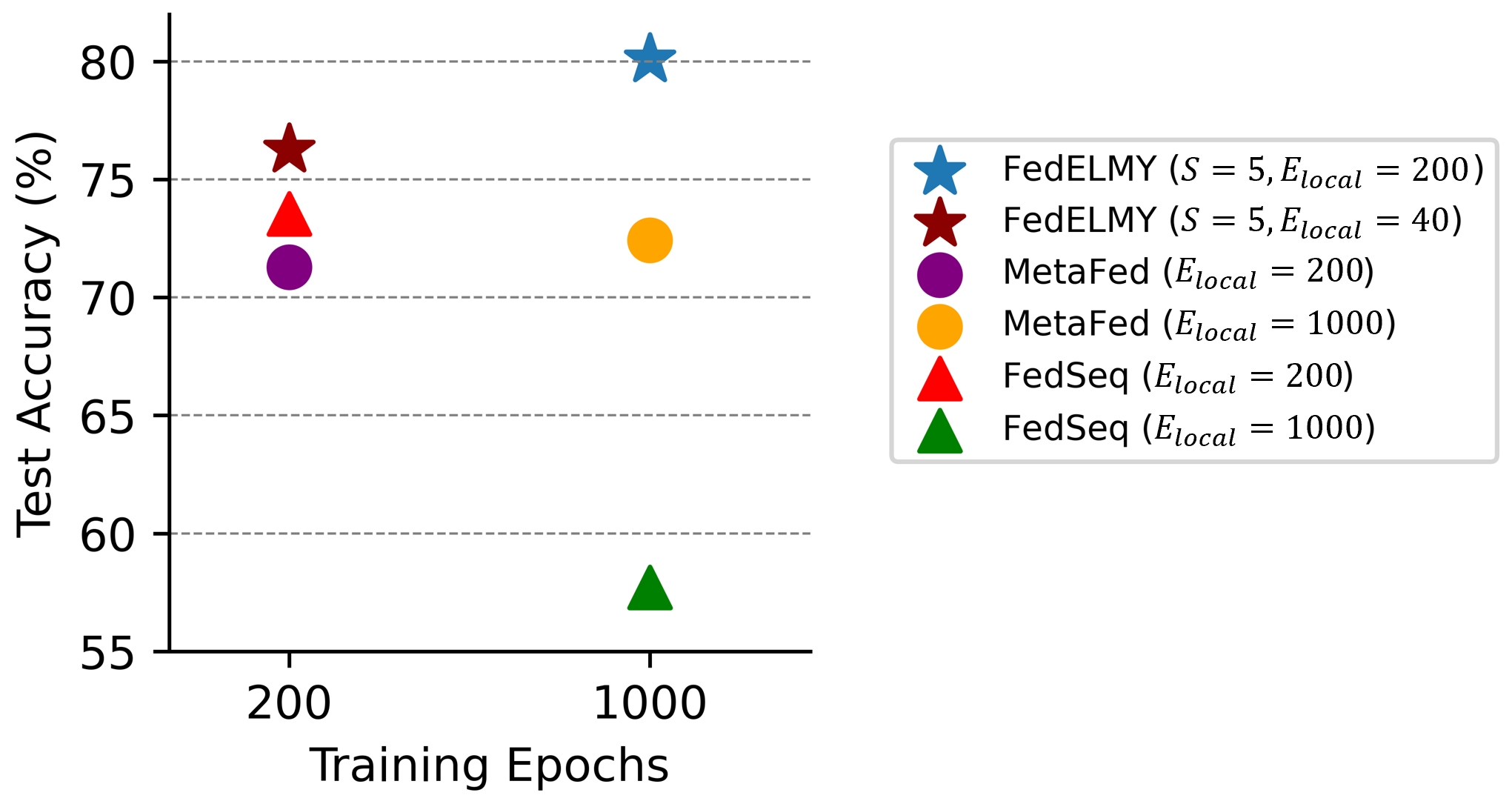}
   \caption{Test accuracy comparison with different computation costs on CIFAR-10 dataset, where \method\ ($S=5, E_{local}=40$) means we will train 5 models on every client, wherein each model undergoes 40 training epochs in our framework; FedSeq ($E_{local}=1000$) denotes that we will train only 1 model for 1000 epochs for every client in FedSeq. }
   \label{fig:baselines}
   \vspace{-0.3cm} 
\end{figure}

\subsubsection{Computation cost.} 


Given that \method\ requires training $S$ models at each client, it involves $S \times E_{local}$ epochs of training per client. Compared to FedSeq, in which each client trains a single model for $E_{local}$ epochs, and MetaFed, which necessitates $2 \times E_{local}$ epochs of training per client, our method appears to demand a higher computational expense. However, \method\ significantly enhances the model performance without increasing the communication cost (Table. \ref{tab:main_results}). Thus, our method strikes a balance (trade-off) between computational cost (training time) and performance gains. 


Meanwhile, we conducted experiments to align the computational costs of different methods for a fair comparison (Fig. \ref{fig:baselines}). First, we align the computation cost of our method to the baselines. By setting \{$S=5, E_{local}=40\}$, \method\ maintains the same computational cost as baselines. As we can see, even under this setting, our method still outperforms the baselines; then, we align the computation cost of the baselines to \method\ by increasing $E_{local}$ to 1000, Fig. \ref{fig:baselines} shows that FedSeq displays worse performance and MetaFed shows negligible variation in its performance. This can be attributed to the phenomenon of overfitting due to excessive local training, which increases the model's generalization error and hence, compromises their performance. Contrastingly, our approach, which equally included $S \times E_{local}$ training rounds, exhibited superior performance. These experiments all validate the efficacy of our model training procedures by the diversity-enhanced mechanism.

\begin{table}[t]
  \centering
  \renewcommand{\arraystretch}{1.1} 
  \caption{Ablation studies on different regularization terms.}
    \begin{tabular}{c|ccc|cc}
    \myhline
    \textbf{Method} &  $\mathcal{M}$ & $d1$  & $d2$ & \textbf{CIFAR10} & \textbf{PACS} \\
    \hline
    MetaFed &       &       &  & 71.29\% & 41.62\% \\
    FedSeq &       &       &  & 73.54\% & 47.31\% \\
    \method & $\checkmark$ &       &  & 78.92\% & 48.11\% \\
    \method & $\checkmark$ & $\checkmark$ &  & 79.62\% & 49.04\% \\
    \method & $\checkmark$ &       & $\checkmark$ & 79.77\% & 48.94\% \\
    \method  & $\checkmark$ & $\checkmark$ & $\checkmark$ & \textbf{80.08\%} & \textbf{49.14\%} \\
    \myhline
    \end{tabular}%
  \label{tab:distance_ablation}%
\end{table}%


\subsection{Ablation study}


In this section, we examine the impact of distance metrics, hyperparameters, and diversity control measures on our method's performance, as well as the influence of client orders.





\subsubsection{Effects of distance terms.} We conducted ablation experiments to verify the effectiveness of the components in our method.  Since the model pool $\mathcal{M}$ is indispensable, we examined the impact of the two distance regularization terms, $d_1$ and $d_2$, referenced in Eq. \ref{eq:d1} and Eq. \ref{eq:d2} respectively. Table \ref{tab:distance_ablation} illustrates the outcomes of respective schemes. As we can see, the introduction of either $d_1$ or $d_2$ independently improves our method's performance compared to solely using the model pool $\mathcal{M}$, and further enhancement is achieved when both distance terms are incorporated. This confirms that integrating both distance terms will strengthen the model's performance. It is worth emphasizing that even in the absence of $d_1$ and $d_2$, relying solely on the model pool $\mathcal{M}$ for training, our approach still outperforms the baselines. This reaffirms the essential paradigm for fostering diversity within the system.

\begin{figure}
  \begin{subfigure}{0.46\linewidth}
  \centering
    \includegraphics[width=\linewidth]{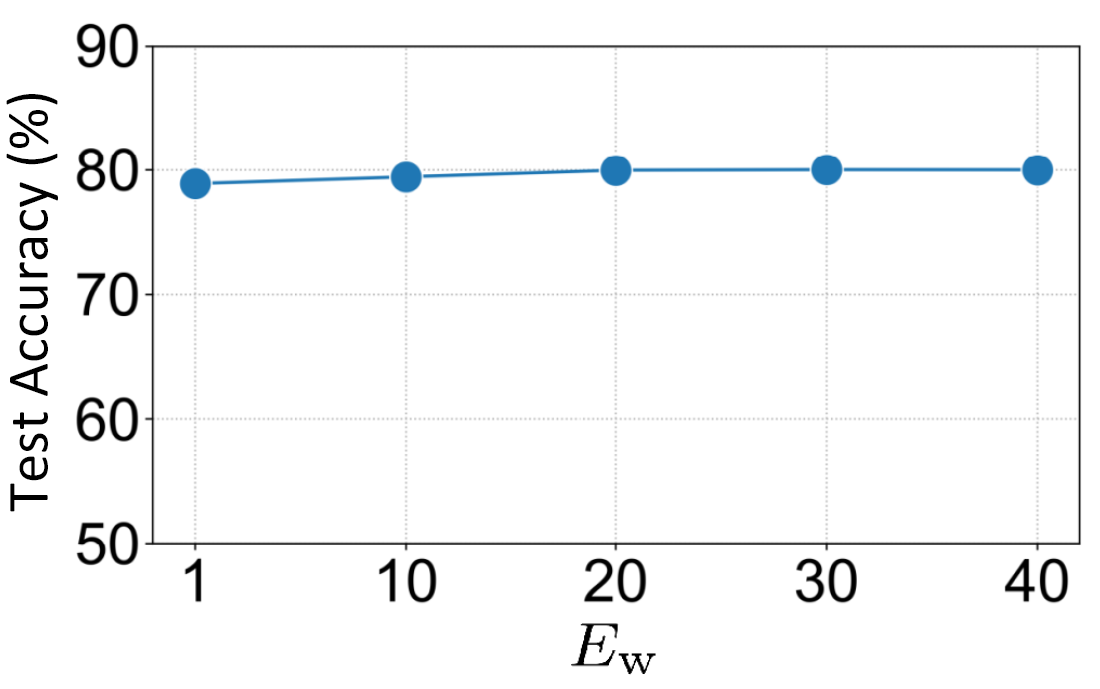}
    \caption{Effect of warm-up epoch $E_{\mathrm{w}}$.}
  \end{subfigure}
  \begin{subfigure}{0.46\linewidth}
  \centering
    \includegraphics[width=\linewidth]{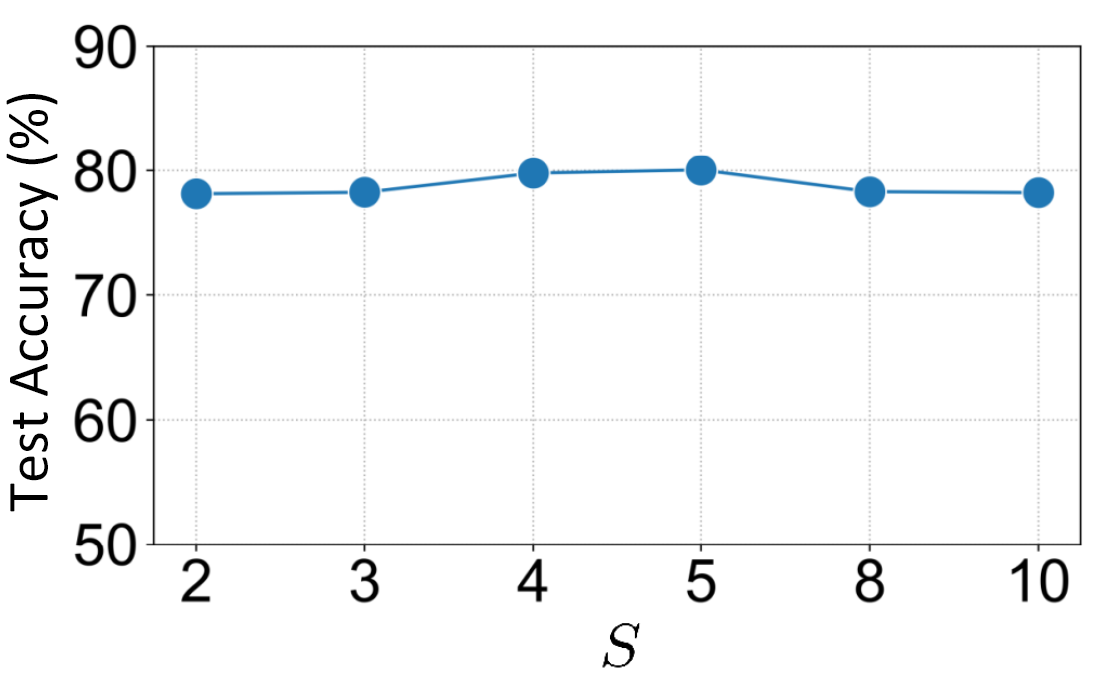}
    \caption{Effect of model quantity $S$.}
  \end{subfigure}
  
  \begin{subfigure}{0.46\linewidth}
  \centering
    \includegraphics[width=\linewidth]{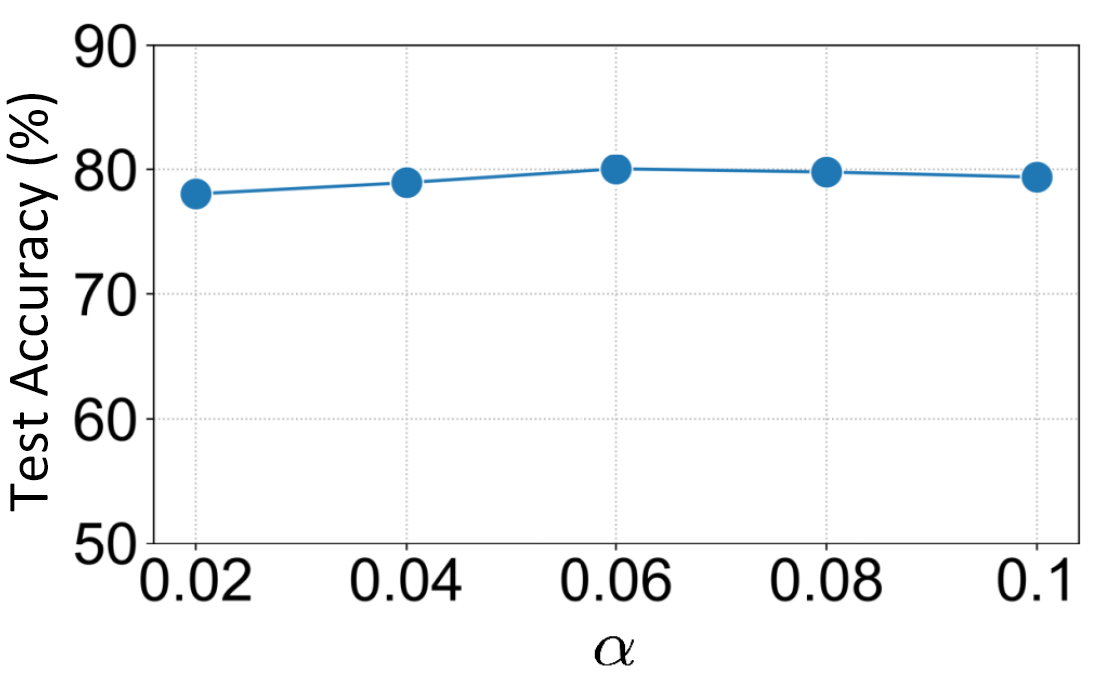}
    \caption{Effect of scale parameter $\alpha$.}
  \end{subfigure}
  \begin{subfigure}{0.46\linewidth}
  \centering
    \includegraphics[width=\linewidth]{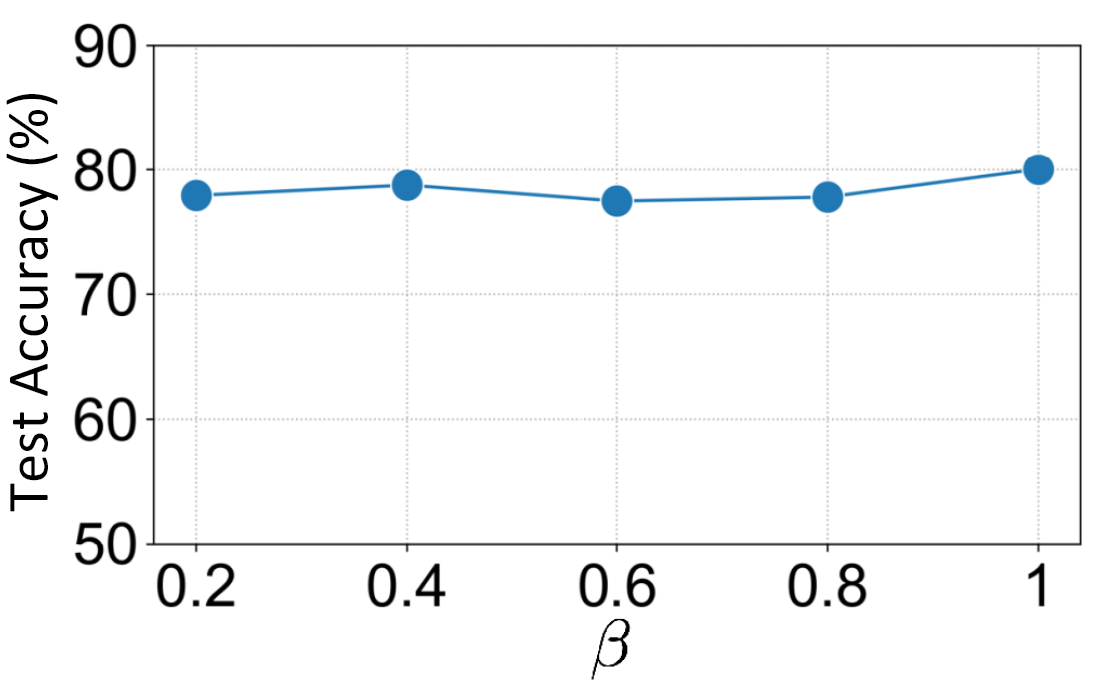}
    \caption{Effect of scale parameter $\beta$.}
  \end{subfigure}
  \caption{Grid search results for CIFAR-10 dataset to investigate the sensitivity of \method\ to various hyperparameters.}
   \label{fig:grid_search}
\end{figure}


\subsubsection{Hyperparameter sensitivity.} We further investigate the sensitivity of our method to different choices of hyperparameters, as shown in Fig. \ref{fig:grid_search}. The optimal combination of these parameters was then employed in subsequent training. It is evident from the figure that regardless of the adopted search strategies, the performance robustness of our method remains relatively undisturbed across all four examined parameters. 
This evidence suggests that the performance of our method is robust to the change of hyperparameters.





\begin{figure*}[t]
  \centering
  \begin{subfigure}[b]{0.32\linewidth} 
    \includegraphics[width=\linewidth]{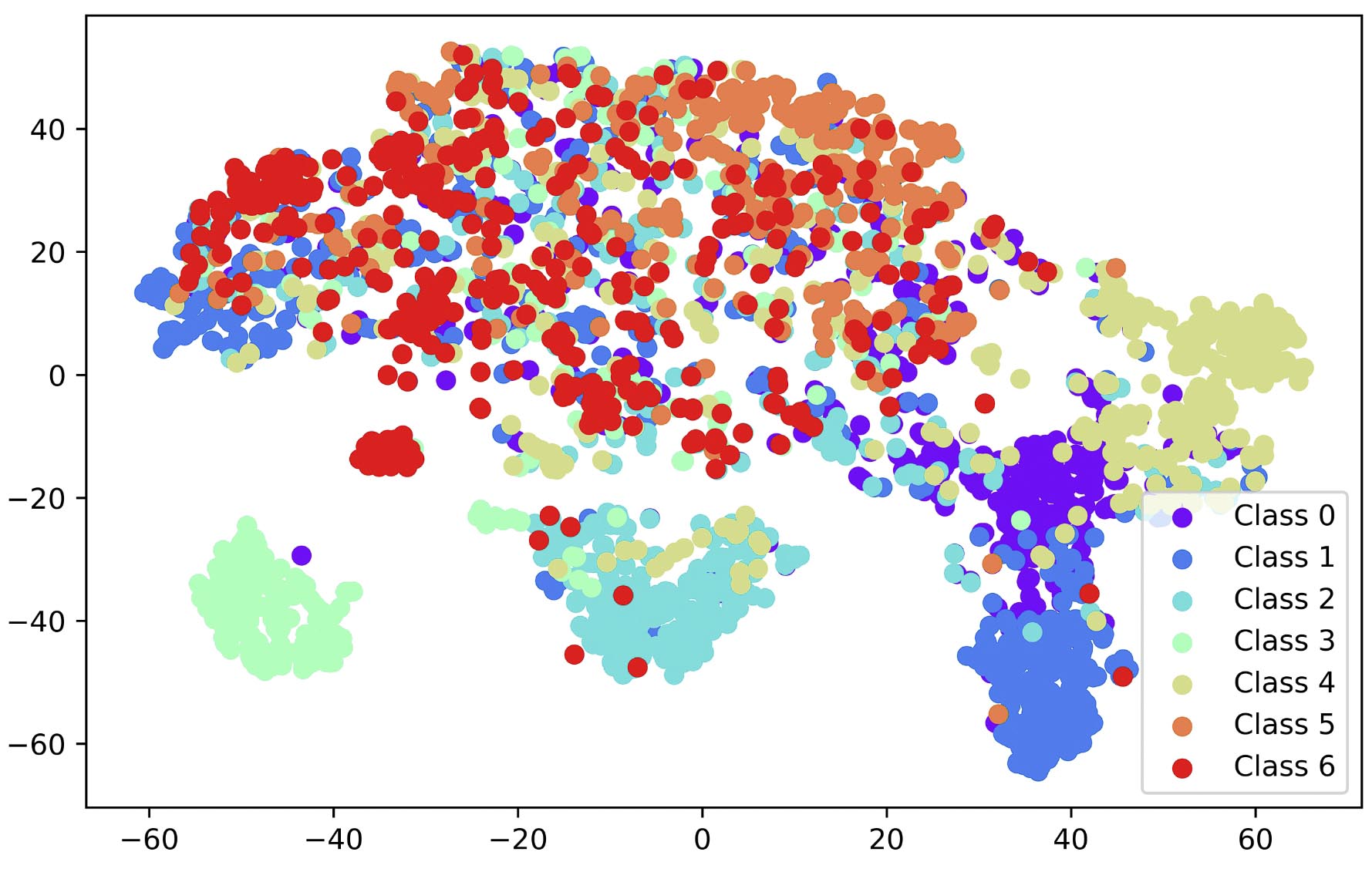}
    \caption{\method}
  \end{subfigure}
  \hfill
  \begin{subfigure}[b]{0.32\linewidth}
    \includegraphics[width=\linewidth]{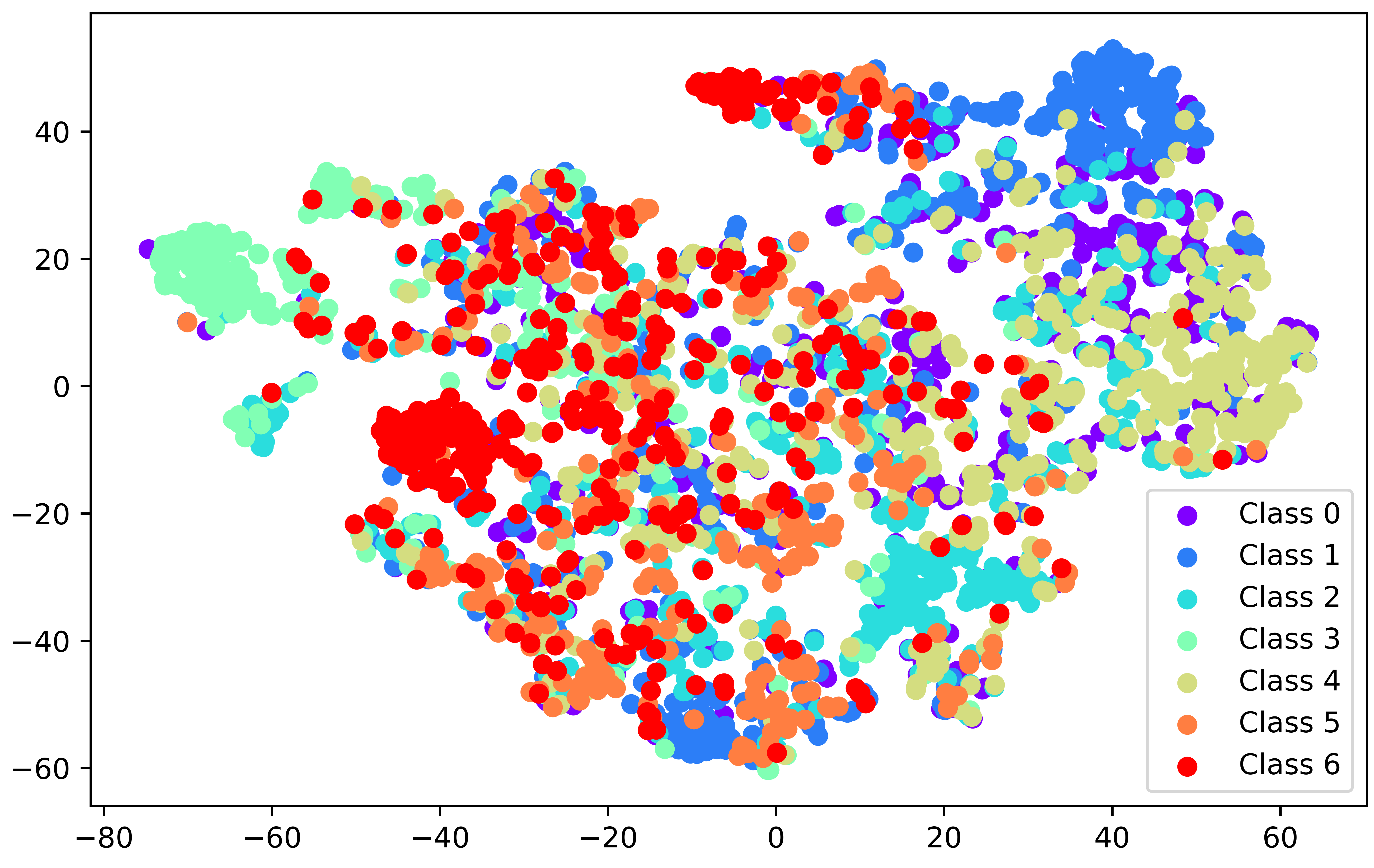}
    \caption{MetaFed}
  \end{subfigure}
  \hfill
  \begin{subfigure}[b]{0.32\linewidth}
    \includegraphics[width=\linewidth]{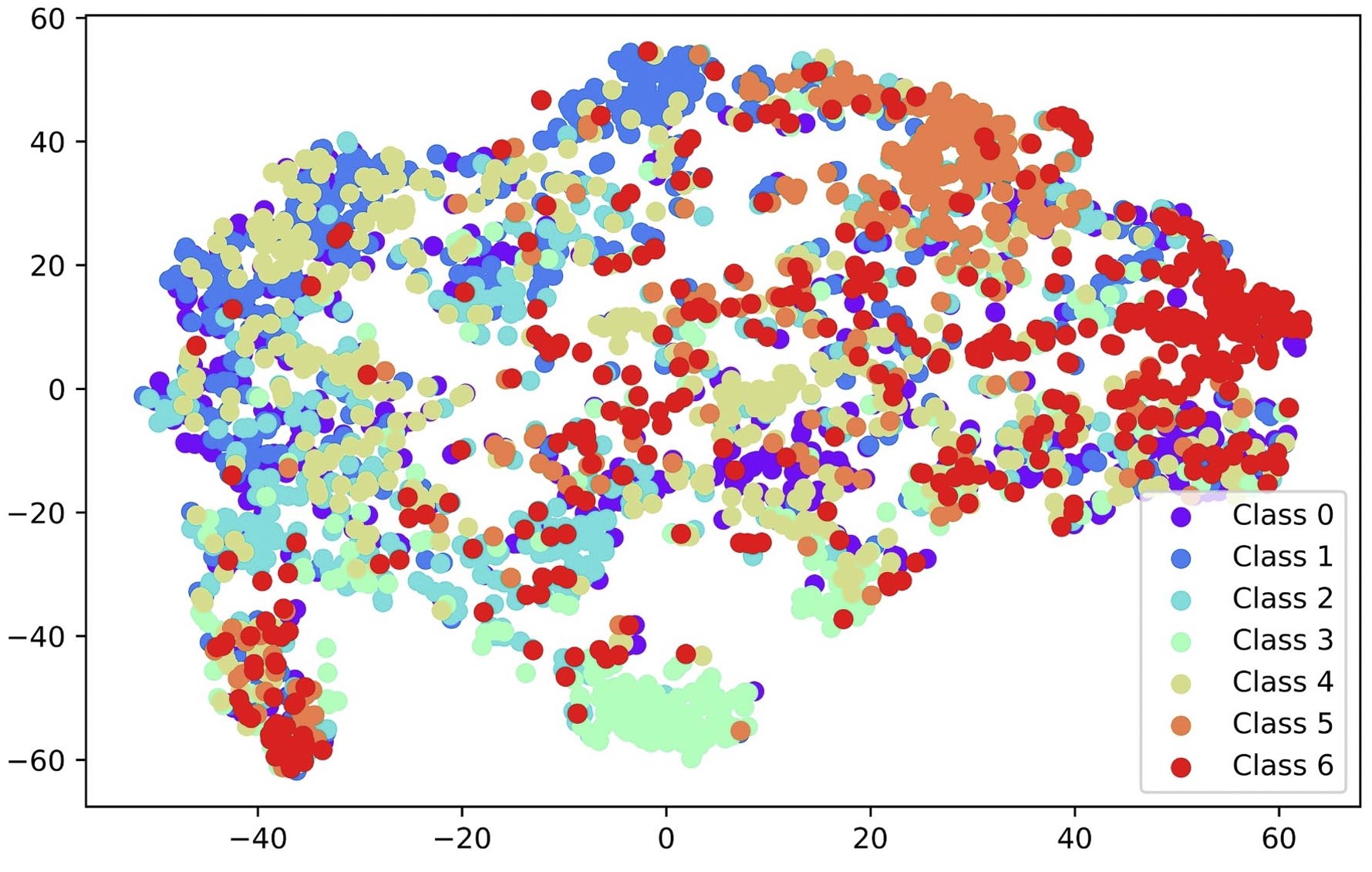}
    \caption{FedSeq}
  \end{subfigure}
  \caption{T-SNE comparison between \method\ and other SFL baselines on PACS dataset with model ResNet-18.}
  \label{fig:tsne}
\end{figure*}

\begin{table}[t]
  \centering
  \renewcommand{\arraystretch}{1.1} 
  \caption{Performance comparison of PACS dataset for different domain training orders, where the order ``PACS'' means we train models by domain order  ``\textit{Photo} (client 1) $\rightarrow$ \textit{Art-Painting} (client 2) $\rightarrow$ \textit{Cartoon} (client 3) $\rightarrow$ \textit{Sketch} (client 4)''.}
    \begin{tabular}{cccccc}
    \myhline
    \textbf{Order} & \textbf{PACS} & \textbf{ACPS} & \textbf{SCPA} & \textbf{CSPA} & \textbf{Average} \\
    \hline
    MetaFed & 41.62\% & 42.75\% & 31.65\% & 40.64\% & 39.17\% \\
    FedSeq & 47.31\% & 45.21\% & 40.96\% & 33.71\% & 41.80\% \\
    \method  & \textbf{49.14\%} & \textbf{46.74\%} & \textbf{43.65\%} & \textbf{41.46\%} & \textbf{45.25\%} \\
    \myhline
    \end{tabular}%
  \label{tab:order}%
\end{table}%


\begin{figure}[t]
  \centering
   \includegraphics[width=.9\linewidth]{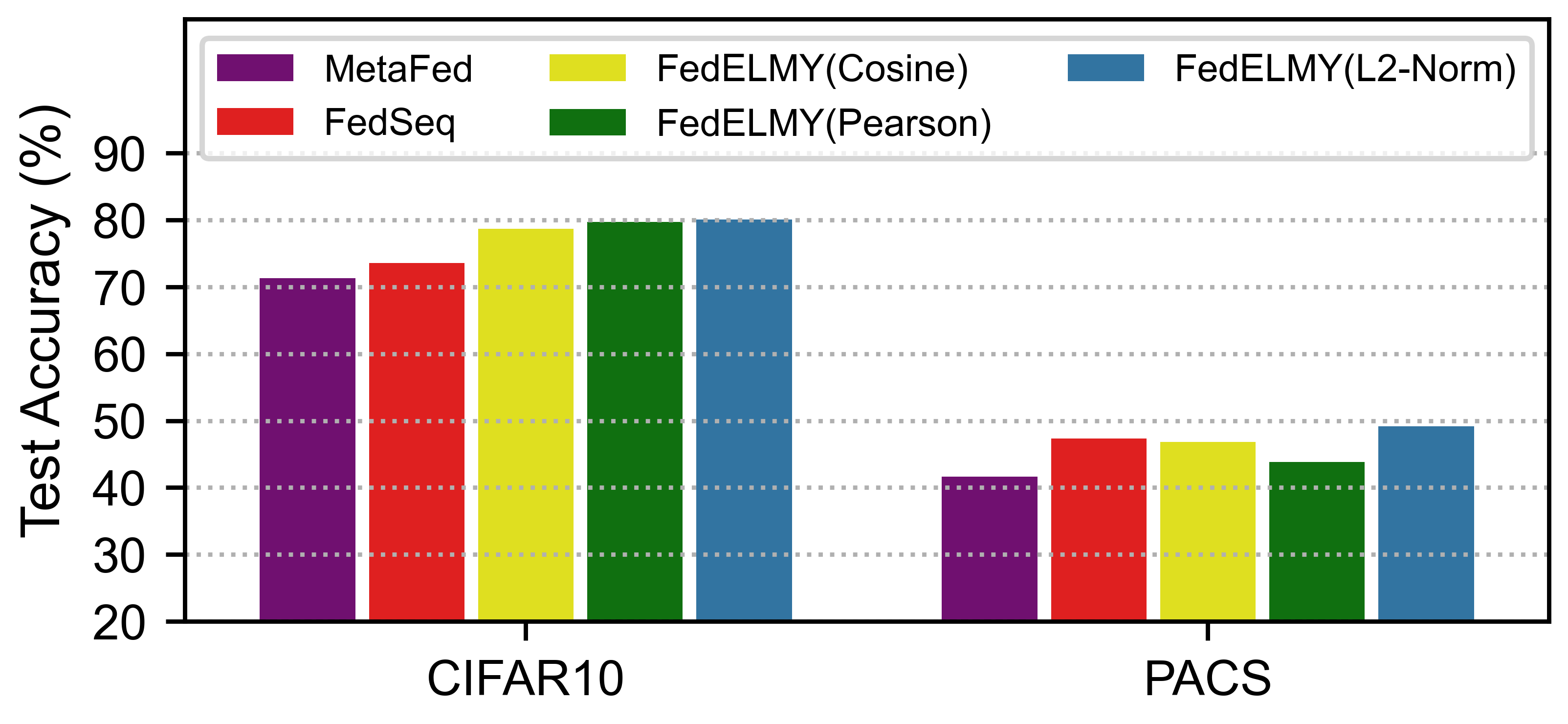}
   \caption{Test accuracy comparison with different diversity control measures, where \method\ (L2-Norm) means we will use the L2-norm (Euclidean) distance as our diversity control measure to train the models.}
   \label{fig:diversity_metrics}
\end{figure}

\begin{figure}[t]
  \centering
  \begin{subfigure}{0.48\linewidth}
  \centering
    \includegraphics[width=\linewidth]{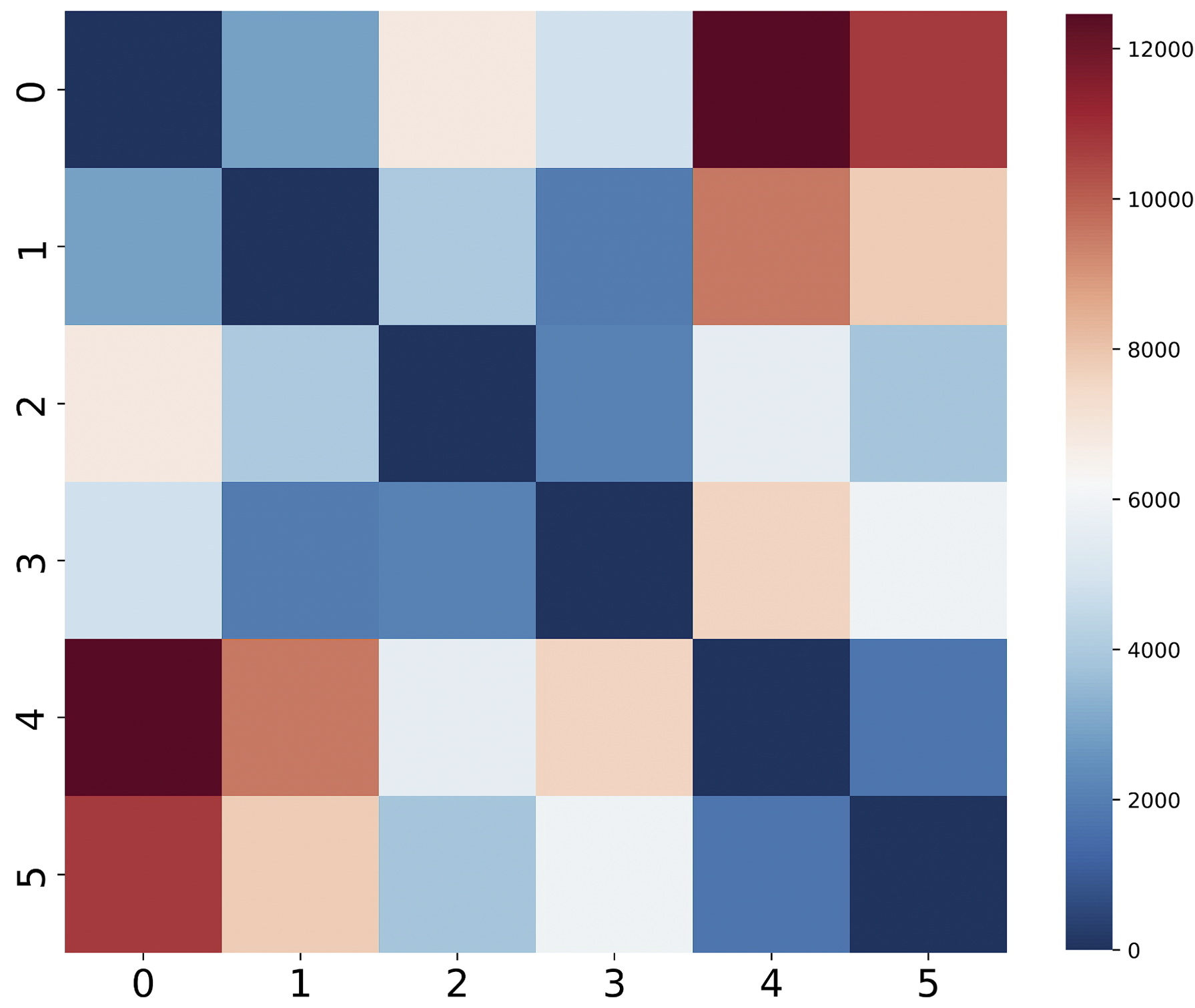}
    \caption{CIFAR-10 ($S=5$)}
  \end{subfigure}
  \hspace{0.1cm}
  \begin{subfigure}{0.48\linewidth}
  \centering
    \includegraphics[width=\linewidth]{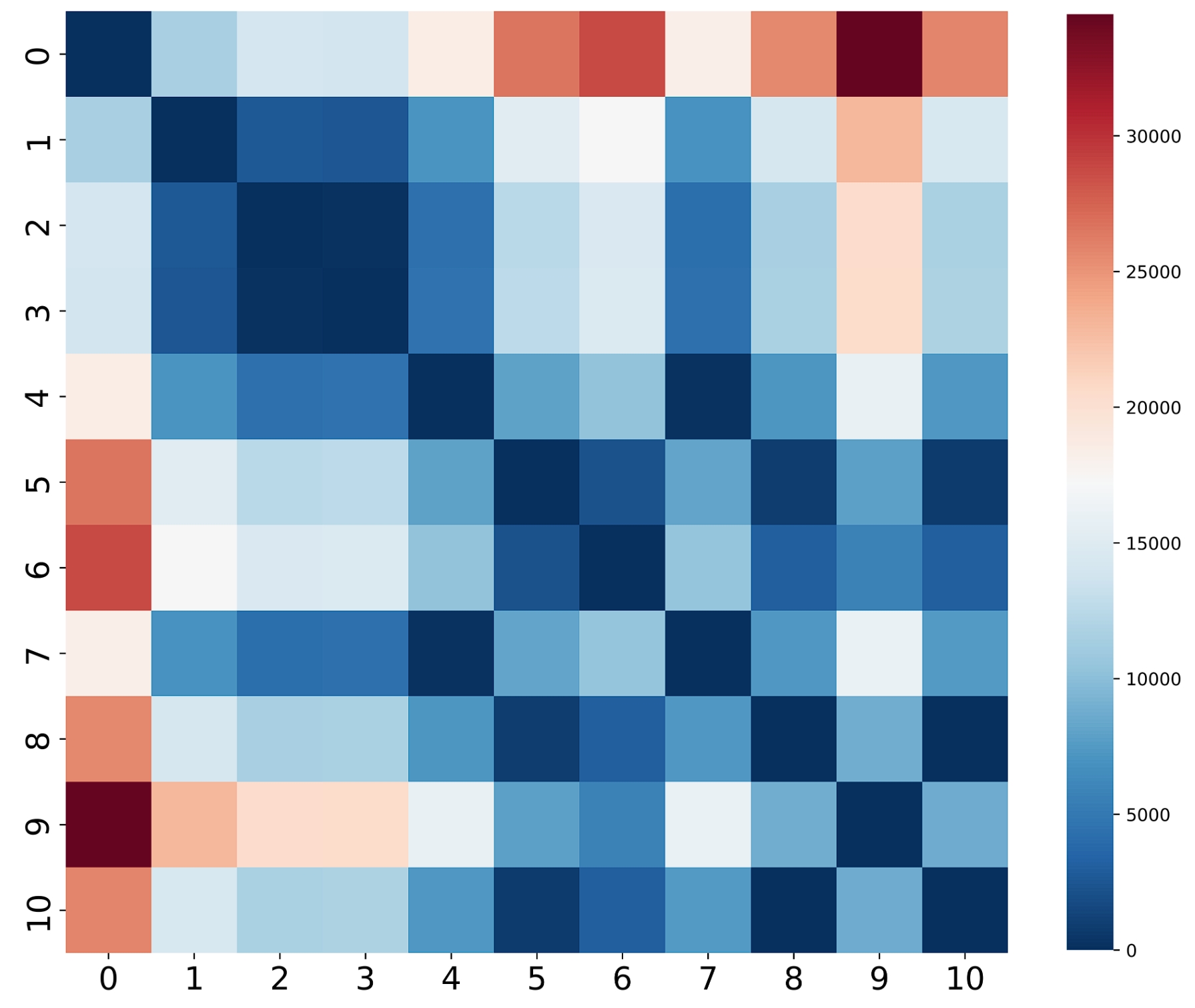}
    \caption{PACS ($S=10$)}
  \end{subfigure}
  \caption{Heatmap of pairwise L2-norm (Euclidean) distances of all trained models within the final client's model pool $\mathcal{M}^N$ with size $S+1$.}
  \label{fig:heatmap}
\end{figure}

\subsubsection{Client order.} To explore the influence of the order in which domains are trained in domain-shift tasks, we applied our method to the PACS dataset with various domain training orders, as depicted in Table \ref{tab:order}. As we can see, irrespective of the order chosen for domain training, our method consistently surpasses the performance of the baseline methods. This signifies our method's robustness in handling diverse domain order training. 

\subsubsection{Diversity control measures.} 
Fig. \ref{fig:diversity_metrics} presents the impact of various diversity control measures applied with different distances. Notably, the L2-norm emerges as the best measure, yet, the remaining measures also exceed the performance of baseline methods in most cases. This superiority of the L2-norm is attributed to its ability to measure the true distance between points directly and precisely, offering a more sensitive and effective constraint on point dispersion in the space, thereby yielding superior results.

\subsection{Case study}


\subsubsection{Classification result visualization} Fig. \ref{fig:tsne} illustrates the t-SNE plots of the feature maps for our method and other SFL baselines on the PACS dataset with model ResNet-18. As we can see, in comparison to MetaFed and FedSeq, \method\ more effectively clusters samples of the same class, indicating a clear separation in the feature space for class 0, 1, 2, 3, and 4---a capability not observed in other baselines. Consequently, by refining the diverse local training process, our approach is adept at learning better feature representations, thereby improving overall model performance.


\subsubsection{Local model diversities.} Fig. \ref{fig:heatmap} illustrates the L2-norm distance matrix for each pair of models within the final client's model pool $\mathcal{M}^N$, after the completion of training. It is evident that the pairwise distances among all trained models in the model pool display a remarkable variation, without an apparent correlation or trend, such as monotonically increasing/decreasing. This confirms the significant diversity among the models within the pool, attesting to the efficacy of our method in fostering model diversity. By strategically enhancing the disparities between models, we can substantively improve the performance of our approach.
\section{Conclusion}




In this paper, we highlight the significance of one-shot sequential federated learning for alleviating the communication burdens of current collaborative machine learning paradigms and address the challenge posed by non-IID data. We present a novel one-shot SFL framework with the local model diversity enhancement strategy to reduce communication costs and effectively improve the global model. In particular, we design the local model pool with two regularization terms as a diversity-enhanced mechanism to improve model performance and mitigate the effect of non-IID data. The effectiveness of our method was demonstrated with superior performance on extensive experiments across several datasets including both label-skew and domain-shift tasks. In the future, we will consider integrating more advanced privacy protection measures, adapting to more federated learning settings, and dealing with real-time data under online learning environments to further enhance the feasibility and scalability of the proposed framework.

\bibliography{main}









\appendix
\clearpage

\LARGE
\noindent\textbf{Appendix}
\normalsize

\section{Experimental Setup}
\label{sec:setup_supp}

\noindent\textbf{Adaptation Details} In an effort to ensure a fair comparison, we adapted the decentralized algorithms DFedAvgM, DFedSAM, and FedSeq to the one-shot setting, and adjusted these methods to select all clients for both training and communication to fit the one-shot setting. This modification means we operate these methods for only one round of communication and select all clients for training and model distribution instead of just a portion, ensuring that all the clients' local data are utilized.



\noindent\textbf{Implementation Details} In all methods, for local model training, we select the model with the highest validation accuracy as the final model. We employed the Adam optimizer (excluding DFedSAM, which utilized the SAM optimizer) with a learning rate of $\eta=5\times 10^{-5}$ applicable for the CIFAR-10, PACS, and Tiny-ImageNet datasets. Due to the relatively smaller size of the Office-Caltech-10 dataset, we adopted the Adam optimizer with a learning rate of $\eta=0.001$. For the MA-Echo method, we used the learning rate of 0.01 for all datasets as specified in their paper. In all cases, the Adam optimizer's weight decay was fixed at $1\times10^{-4}$. As for the other hyperparameters of the baselines, we adhered to the values suggested in their papers, which were kept constant and chosen based on optimal validation accuracy. We applied the optimal hyperparameters obtained through the grid-search strategy for our method. Specifically, we set $E_{\mathrm{w}}$ to $30$ for the CIFAR-10 dataset, $20$ for all other datasets; we set $S$ to $5$ for the CIFAR-10 dataset, $10$ for the PACS and Office-Caltech-10 dataset, and $3$ for the Tiny-ImageNet dataset; we set $\alpha$ to $0.06$ for the CIFAR-10 dataset, $1$ for the PACS and Tiny-ImageNet dataset, $0.001$ for the Office-Caltech-10 dataset; we set $\beta$ to $1$ for the CIFAR-10, PACS and Tiny-ImageNet dataset, $0.001$ for the Office-Caltech-10 dataset. During the local training phase of every client, in addition to utilizing scale hyperparameters $\alpha$ and $\beta$, our method also included control over the magnitudes of the distance parameters $d_1$ and $d_2$ to one order of magnitude smaller than the original loss $\ell$, which was achieved through the logarithmic scaling. For instance, assuming the initial loss $\ell$ is 6.02 and the original distance $d_1$ is 45, we would first normalize $d_1$ to 0.45. Following this, we multiply this normalized distance by the scaling factor $\alpha$ and subsequently integrate it into the comprehensive loss function $\mathcal{L}$. This calibration ensures the original loss maintains primacy in influencing the training process, while the distance regularization terms play an ancillary role, thereby preventing their dominance over the central objective.


\noindent\textbf{Environment} All our experiments were conducted on a single machine with 1TB RAM and 256-core AMD EPYC 7742 64-Core Processor @ 3.4GHz CPU. The GPU we used is NVIDIA A100 SXM4 with 40GB memory. The software environment settings are: Python 3.7.4, PyTorch 1.13.1 with CUDA 11.6 on Ubuntu 20.04.4 LTS. All the experimental results are the average over three trials.



\section{Few-Shot Adaptation}
\label{sec:ring}
Alg. \ref{alg:ring_method} illustrates the implementation details and overview of our method in few-shot scenarios. In this scenario, after the local training of the final client $N$,  it will send its averaged model $m_{avg}^N$ to the first client $1$ to start a new cycle of model training, this process will be repeated for multiple rounds. After completing $r = T$ iterations, we regard $m_{avg}^N$ as the final global model $m_{final}$.


\begin{algorithm}[t]
    \caption{Few-Shot Adaptation of \method}
    \label{alg:ring_method}
    \KwIn{Local datasets $\mathcal{D} = \{D_i\}_{i=1}^{N}$, warm-up epoch $E_{\mathrm{w}}$, learning rate $\eta$, number of local iterations $E_{local}$, model number to be trained per client $S$, scale hyperparameters $\alpha$, $\beta$}
    \KwOut{The final model $m_{final}$}
    
    \textbf{Initialization:} For client 1, warm up a randomly initialized model $m_{avg}^0$ for $E_{\mathrm{w}}$ epochs  
    
        \For{shot $r= 1:T$}{
        \For{client $i= 1:N$ }{
        Receives $m_{avg}^{i-1}$ from client $i-1$ (for client $1$ from shot $r>1$, receives from client $N$)

        \textcolor{lightblue}{// Initialize model pool $\mathcal{M}^i$ for client $i$}

        $\mathcal{M}^i = \{m_0^i\}$ with $m_0^i \leftarrow$ $m_{avg}^{i-1}$
        
        \For{$j=1:S$}{
        
        \textcolor{lightblue}{// Initialize $m_j^i$}
        
        $m_j^i \leftarrow \frac{1}{|\mathcal{M}^i|}\sum_{t=0}^{|\mathcal{M}^i|-1} m_t^i$

        \textcolor{lightblue}{// Local training for $m_j^i$}
 
        \For{$k=1:E_{local}$}{
        
   $\mathcal{L}(m_j^i)\leftarrow\ell(m_j^i; D_i) - \alpha \cdot d_1 + \beta \cdot d_2$

   $m_j^i \leftarrow m_j^i - \eta \nabla_m \mathcal{L}(m_j^i)$
        }

        $\mathcal{M}^i \leftarrow \mathcal{M}^i \cup \{m_j^i\}$
        
        }
        $m_{avg}^i \leftarrow\frac{1}{|\mathcal{M}^i|}\sum_{t=0}^{|\mathcal{M}^i|-1}m_t^i$
        
        Sends $m_{avg}^i$ to client $i+1$ (for client $N$, sends to client $1$)

        }
        
        }
        \textcolor{lightblue}{// For the final client $i=N$, outputs model $m_{final}$}
        
        $m_{final} \leftarrow m_{avg}^N$
\end{algorithm}

\section{Adaptation to Decentralized Parallel Federated Learning}
\label{sec:mesh}

\begin{algorithm}[t]
    \caption{\method\ for Decentralized PFL}
    \label{alg:mesh_method}
    \KwIn{Local datasets $\mathcal{D} = \{D_i\}_{i=1}^{N}$, warm-up epoch $E_{\mathrm{w}}$, learning rate $\eta$, number of local iterations $E_{local}$, model number to be trained per client $S$, scale hyperparameters $\alpha$, $\beta$}
    \KwOut{The final model $m_{final}$}
    
    \textbf{Initialization:} For every client $i$, warm up a randomly initialized model $m_0^i$ for $E_{\rm{w}}$ epochs



        
        \For{client $i= 1:N$ \textbf{in parallel} }{
        

        \textcolor{lightblue}{// Initialize model pool $\mathcal{M}^i$ for client $i$}

        $\mathcal{M}^i = \{m_0^i\}$
        
        \For{$j=1:S$}{
        
        \textcolor{lightblue}{// Initialize $m_j^i$}
        
        $m_j^i \leftarrow \frac{1}{|\mathcal{M}^i|}\sum_{t=0}^{|\mathcal{M}^i|-1} m_t^i$

        \textcolor{lightblue}{// Local training for $m_j^i$}
 
        \For{$k=1:E_{local}$}{

        
        


   
   
   $\mathcal{L}(m_j^i)\leftarrow\ell(m_j^i; D_i) - \alpha \cdot d_1 + \beta \cdot d_2$

   $m_j^i \leftarrow m_j^i - \eta \nabla_m \mathcal{L}(m_j^i)$
        }

        $\mathcal{M}^i \leftarrow \mathcal{M}^i \cup \{m_j^i\}$
        
        }
        $m_{avg}^i \leftarrow\frac{1}{|\mathcal{M}^i|}\sum_{t=0}^{|\mathcal{M}^i|-1}m_t^i$
        
        Sends $m_{avg}^i$ to all other clients $j=1:N$ , ${j} \neq {i}$
        }

        

        \textcolor{lightblue}{// For any client $i=1:N$, output model $m_{final}$}
        
        $m_{final} \leftarrow \frac{1}{N}\sum_{i=1}^{N}m_{avg}^i$
\end{algorithm}

Our method can also be adapted to the decentralized Parallel Federated Learning (PFL) setting. Alg. \ref{alg:mesh_method} demonstrates the implementation details of our method under the decentralized PFL environment. As we can see, under this setting, clients will train their models concurrently (in parallel). During the training process, each client begins with a randomly initialized model $m_0^i$. Upon completion of local training, each client will simultaneously transmit the locally averaged model $m_{avg}^i$ to all other clients. 
Once a client $i$ receives all models from every neighbor, it will calculate an aggregated model by averaging its own averaged model $m_{avg}^i$ with those received averaged models as the final model $m_{final}$. 
Please note that our framework can also be adjusted for a centralized PFL environment with a central server. In this setup, each client will send its averaged model $m_{avg}^i$ to the central server. The overall performance will be consistent with that of the decentralized PFL adaptation since the server will finally average all models in a manner identical to the decentralized setting. Experimental results of such an adaptation are shown in Sec. \ref{sec:exps_supp}.


\section{Additional Experiments}

In the remaining parts, unless otherwise specified, we set the local training epoch $E_{local}$ to 200, use the ResNet-18 model structure and follow the Dirichlet distribution $Dir(0.5)$ for the label-skew tasks in our experiments. 

\label{sec:exps_supp}


\begin{table}[t]
  \centering
  \renewcommand{\arraystretch}{1.1} 
  \caption{Test accuracy (\%) comparison on more public datasets, including both label-skew and domain-shift datasets.}
      \scalebox{0.92}{
    \begin{tabular}{c|cc|cc}
    \myhline
    Distrubution & \multicolumn{2}{c|}{Label-Skew}  & \multicolumn{2}{c}{Domain-Shift} \\
    \hline
    Dataset & \textbf{MNIST} & \textbf{SVHN} & \textbf{Office31} & \makecell{\textbf{Office-}\\ \textbf{Home}}\\
    \hline
    DFedAvgM &   11.35$\pm$4.01    &    18.60$\pm$5.66   &    3.47$\pm$1.11   &  1.71$\pm$0.23 \\
    DFedSAM &     12.53$\pm$3.93  &   22.97$\pm$4.98    &     3.49$\pm$1.29  &  1.25$\pm$0.32 \\
    FedOV &    73.10$\pm$0.33   &    37.47$\pm$1.48   &    3.39$\pm$0.46   &  2.31$\pm$0.15 \\
    DENSE &    95.88$\pm$1.59   &    73.76$\pm$1.25   &     3.15$\pm$0.58  &  1.68$\pm$0.17 \\
    \hline
    MetaFed &  98.07$\pm$0.39   &   76.66$\pm$1.78    &    2.97$\pm$0.43   &  2.40$\pm$0.50  \\
    FedSeq &    98.31$\pm$0.65   &   76.18$\pm$2.45    &   3.46$\pm$0.22    &  2.47$\pm$0.29  \\
    \method &  \textbf{98.91$\pm$0.23}    &    \textbf{79.55$\pm$4.12}   &    \textbf{3.57$\pm$0.09}   & \textbf{2.70$\pm$0.13} \\
    \hline
    \myhline
    \end{tabular}%
    \label{tab:moredatasets}
    }
\end{table}%

\vspace{3mm}
\noindent\textit{D.1 Comparison on more datasets.} We compare our method to other baselines on four more public datasets, including both label-skew and domain-shift datasets to enrich our experimental results. As shown in Table \ref{tab:moredatasets}, our method consistently outperforms all other baselines on both label-skew and domain-shift scenarios, which further validates the effectiveness of our approach.

\begin{figure}
  \begin{subfigure}{0.46\linewidth}
  \centering
    \includegraphics[width=\linewidth]{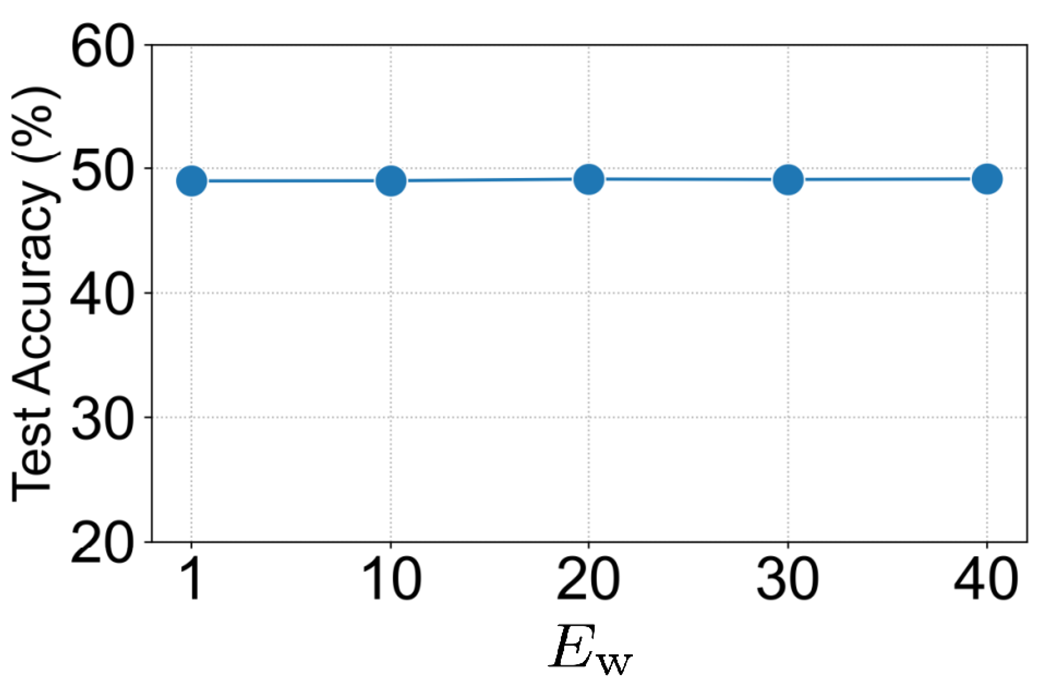}
    \caption{Effect of warm-up epoch $E_{\mathrm{w}}$.}
  \end{subfigure}
  \begin{subfigure}{0.46\linewidth}
  \centering
    \includegraphics[width=\linewidth]{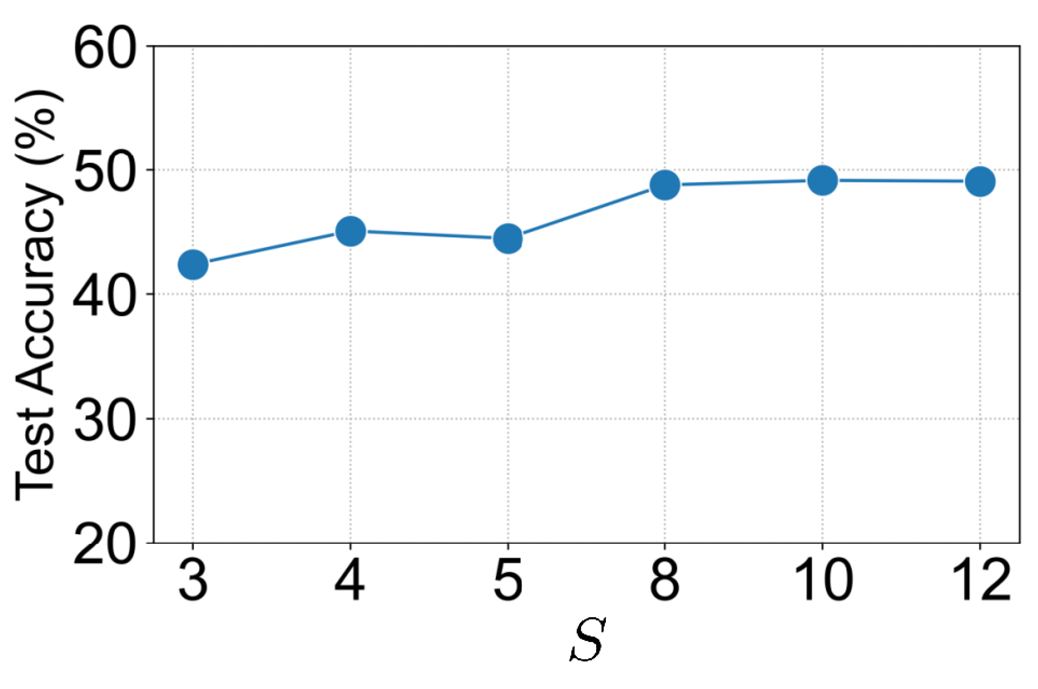}
    \caption{Effect of model quantity $S$.}
  \end{subfigure}
  
  \begin{subfigure}{0.46\linewidth}
  \centering
    \includegraphics[width=\linewidth]{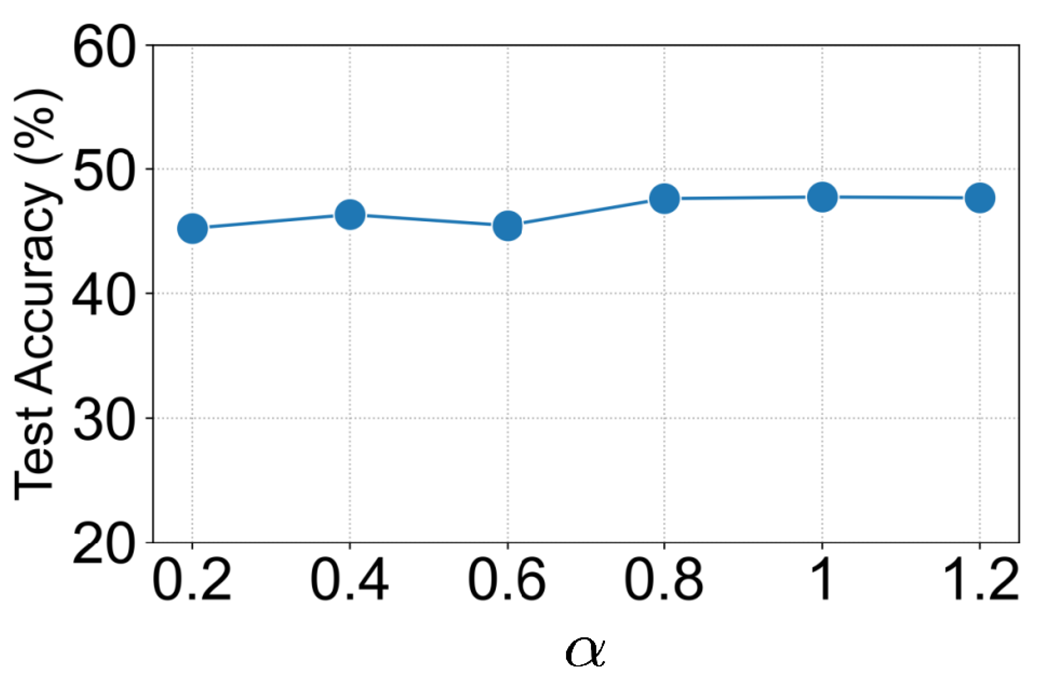}
    \caption{Effect of scale parameter $\alpha$.}
  \end{subfigure}
  \begin{subfigure}{0.46\linewidth}
  \centering
    \includegraphics[width=\linewidth]{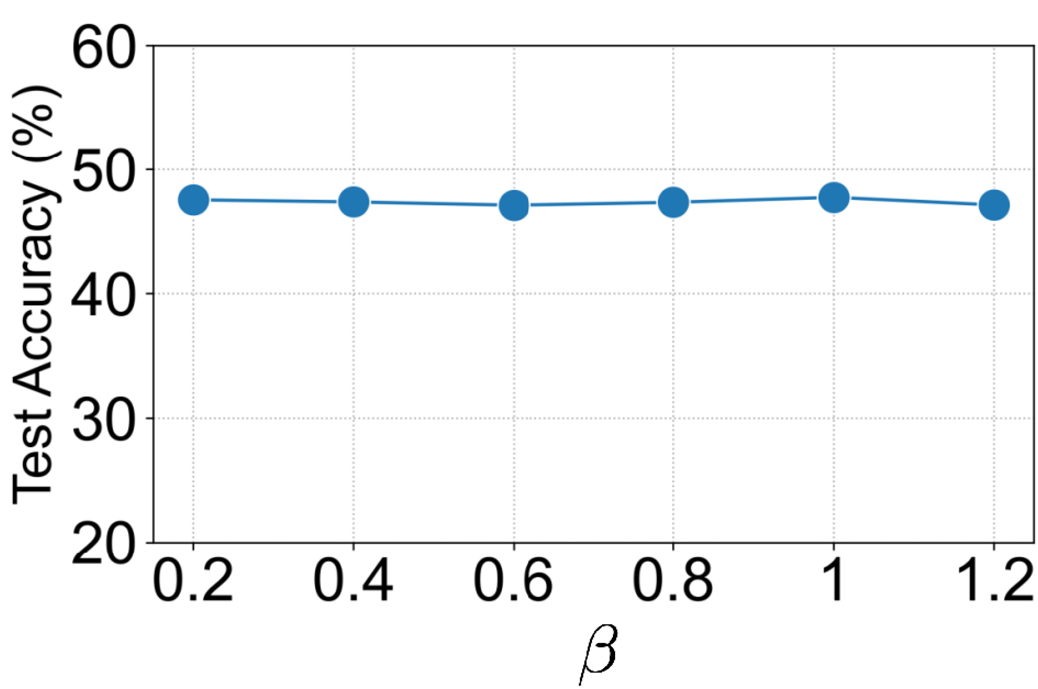}
    \caption{Effect of scale parameter $\beta$.}
  \end{subfigure}
  \caption{Grid search results for PACS dataset to investigate the sensitivity of our method to different hyperparameters.}
   \label{fig:grid_search_pacs}
\end{figure}

\vspace{3mm}
\noindent\textit{D.2 Hyperparameter sensitivity.} We investigate the sensitivity of our method to different choices of hyperparameters for the PACS dataset as a supplement, as shown in Fig. \ref{fig:grid_search_pacs}. The findings show that our method maintains consistent robustness irrespective of the search strategies implemented, as observed across all four evaluated hyperparameters (when $S$ grew to around 10, the performance started to become stable). Such evidence underscores the method's limited reliance on hyperparameter selection.

\begin{figure}[t]
  \centering
   \includegraphics[width=\linewidth]{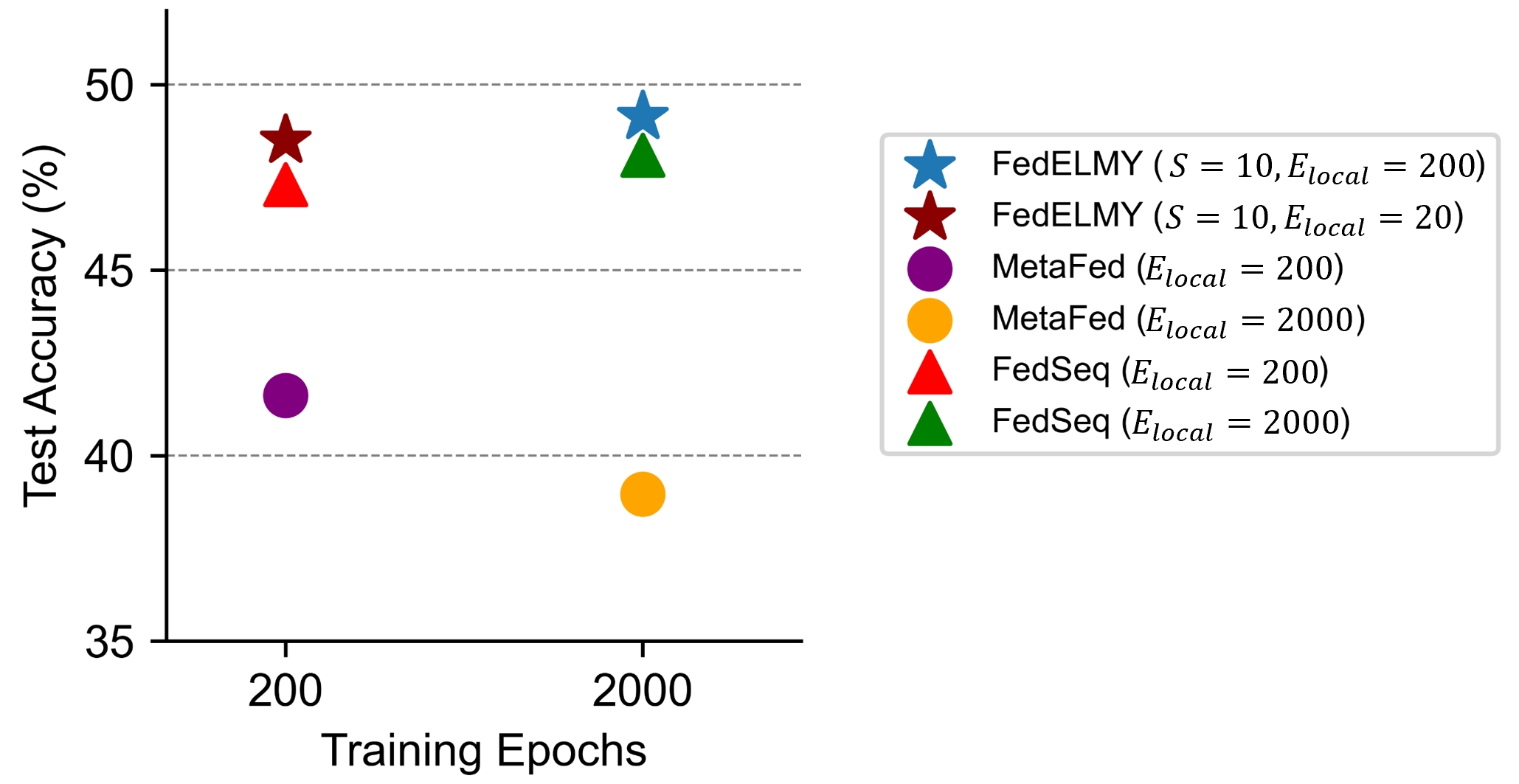}
   \caption{Test accuracy comparison with different computation costs on PACS dataset.}
   \label{fig:computation_cost_pacs}
\end{figure}

\vspace{3mm}
\noindent\textit{D.3 Impact of computation cost.} We show another experiment about the effect of different computation costs on the PACS dataset with Resnet-18 structure. As we can see from Fig. \ref{fig:computation_cost_pacs}, our method can still get the best performance under different computation cost settings, which validates the efficacy of our model training procedures by the diversity-enhanced mechanism.

\begin{table*}[t]
    \centering
    \caption{Test accuracy (\%) comparison of our \method\ method to other baselines on different numbers of clients $N$. For the domain-shift tasks like the PACS dataset, the number of clients $N=8$ means we split the whole dataset into 8 parts and allocate them to the clients following the order of ``\textit{Photo} (client 1) $\rightarrow$ \textit{Art-Painting} (client 2) $\rightarrow$ \textit{Cartoon} (client 3) $\rightarrow$ \textit{Sketch} (client 4) $\rightarrow$ \textit{Photo} (client 5) $\rightarrow$ \textit{Art-Painting} (client 6) $\rightarrow$ \textit{Cartoon} (client 7) $\rightarrow$ \textit{Sketch} (client 8)" for subsequent training.}

\begin{tabular}{c|cccc|cccc|ccc|ccc}
\myhline
Distribution & \multicolumn{8}{c|}{Label-Skew} & \multicolumn{6}{c}{Domain-Shift} \\
\hline
Dataset & \multicolumn{4}{c|}{CIFAR-10} & \multicolumn{4}{c|}{Tiny-ImageNet} & \multicolumn{3}{c|}{PACS} & \multicolumn{3}{c}{Office-Caltech-10} \\
\hline
$N$ & 5   & 20   & 50   & 100   & 5   & 20   & 50   & 100   & 8   & 20 & 40 & 8 & 20 & 40 \\
\hline

DFedAvgM  & 21.27   & 18.22   & 19.50 &     24.48   & 2.64  & 1.96  & 1.99    &    3.09    & 17.55  & 16.37  & 16.62  & 10.65  & 9.73   & 8.94           \\
DFedSAM   & 30.62   & 18.75   & 10.00 &   10.01     & 4.34  & 2.67  & 0.64    &   0.51     & 16.56  & 16.48  & 16.58  & 14.88  & 11.58  & 11.52          \\
FedOV     & 41.49   & 30.93   & 31.93 &    38.42    & 1.33  & 1.23  & 1.11    &     1.08   & 18.59  & 17.73  & 17.32  & 15.78  & 10.56  & 9.34           \\
DENSE     & 67.54   & 52.77   & 48.38 &    19.04    & 3.88  & 3.03  & 3.12    &   3.01     & 19.18  & 9.24   & 14.53  & 20.72  & 9.52   & 11.07          \\
\hline
MetaFed   &  72.53  &  57.73  &   47.47    &  32.13  &     31.44   &    20.46   &   16.16    &    13.04   &   27.33    &    20.35   &  16.76   &   19.43  &    11.33 & 11.45   \\
FedSeq       & 73.84   & 64.05   & 53.62 &    35.82    & 31.97  & 23.38 & 14.92  &    15.27    & 33.01  & 17.29  & 12.20  & 18.40  & 12.23  & 9.78           \\
\method   & \textbf{80.99} & \textbf{68.18} & \textbf{64.31} &     \textbf{40.01}   & \textbf{38.43} & \textbf{29.52} & \textbf{19.93} &    \textbf{20.83}    & \textbf{36.10} & \textbf{25.19} & \textbf{23.60} & \textbf{24.45} & \textbf{12.48} & \textbf{12.21} \\

\myhline
\end{tabular}
    \label{tab:N}
\end{table*}%

\begin{table}[t]
  \centering
  \renewcommand{\arraystretch}{1.1} 
  \caption{Test accuracy (\%) comparison of our \method\ method to other baselines with the CNN model structure.}
\begin{tabular}{ccccc}
\myhline
\textbf{Dataset} & \textbf{CIFAR-10} & \makecell{\textbf{Tiny-}\\ \textbf{ImageNet}} & \textbf{PACS} & \makecell{\textbf{Office-}\\ \textbf{Caltech-10}} \\
\hline
DFedAvgM   & 15.08   & 3.56   & 20.56   & 22.84   \\ 
DFedSAM    & 13.25   & 0.49   & 15.79   & 9.20 \\ 
FedOV      & 53.49   & 0.69   & 12.17   & 22.65   \\ 
DENSE      & 61.42   & 6.72   & 27.16   & 34.88   \\ \hline
MetaFed & 56.11 & 6.68 & 28.46 & 29.99 \\
FedSeq     & 66.08   & 14.49  & 38.36   & 38.22   \\ 
\method       & \textbf{69.49}      & \textbf{17.05}     & \textbf{39.89}      & \textbf{39.77}      \\ \myhline
\end{tabular}
  \label{tab:cnn}%
\end{table}%

\vspace{3mm} 
\noindent\textit{D.4 Impact of client numbers.} We assessed the performance of various methods across all datasets by changing the number of clients $N$, as shown in Table \ref{tab:N}. It is evident that there is a general trend toward diminishing accuracy for all methods as the client number $N$ increases, aligning with findings reported in other literature on traditional federated learning. However, it is noteworthy that despite the influence of client quantity on one-shot federated learning, our approach consistently surpasses other baseline methods in performance. To conclude, these experiments consistently confirm the effectiveness and robustness of our proposed method to deal with one-shot sequential federated learning.

\vspace{3mm}
\noindent \textit{D.5 Impact of model structure.} To evaluate the scalability of our approach, we modified the model structure to a three-layer convolutional neural network (CNN) and replicated the corresponding experiments across all datasets. The results of these tests, which compare the test accuracy of our method to various baseline methods under different scenarios, are presented in Table~\ref{tab:cnn}, consistent with the primary analyses reported in the main paper. These findings affirm the robustness and scalability of our proposed method.


\begin{table}[t]
    \centering
    \caption{Test accuracy (\%) comparison of \method\ to other baselines for different Dirichlet distributions.}
    \begin{tabular}{c|ccc|ccc}
    \myhline
    Dataset & \multicolumn{3}{c|}{CIFAR-10} & \multicolumn{3}{c}{Tiny-ImageNet} \\
    \hline
    $Dir(\cdot)$ & 0.1   & 0.3 & 0.5  & 0.1   & 0.3 & 0.5 \\
    \hline
    DFedAvgM  & 10.33  & 16.14    &      18.50       & 1.07    & 1.68   &         2.04    \\
    DFedSAM  & 15.62  & 21.84     &     19.48       & 1.74    & 2.80    &        3.41     \\
    FedOV    & 25.92  & 31.93     &      45.69      & 0.94    & 1.23    &        1.59     \\
    DENSE    & 26.53  & 61.73    &     64.33       & 1.94    & 2.37   &     1.48         \\
    \hline
    MetaFed   &  27.18  &   57.24  &    71.29    &     9.96   &   19.45   &        24.53    \\
    FedSeq   & 27.41  & 57.91   &    73.54   & 15.00      & 22.23    &       24.74    \\
    \method  & \textbf{29.13}   & \textbf{66.32}  &  \textbf{80.08}    & \textbf{16.76}   & \textbf{26.85} &   \textbf{30.49}  \\
    \myhline
    \end{tabular}%
    \label{tab:beta}
\end{table}%

\vspace{3mm}
\noindent\textit{D.6 Impact of data distribution.} To investigate the label-skew scenarios using the CIFAR-10 and Tiny-ImageNet datasets, we performed experiments across varying levels of skew by manipulating the Dirichlet distribution, denoted as $Dir(\cdot)$. The results, detailed in Table \ref{tab:beta}, reveal that our approach outperforms other competing methods in all scenarios, which demonstrates considerable advantages of our method, particularly in terms of scalability and privacy preservation, marking it as a more robust solution in these contexts.


\begin{table}[t]
  \centering
  \caption{Test accuracy (\%, mean$\pm$std) comparison of our method under the decentralized PFL setting to other similar baselines with the Resnet-18 model structure.}
      \scalebox{0.9}{
    \begin{tabular}{ccccc}
    \myhline
    \textbf{Dataset} & \textbf{CIFAR-10} & \makecell{\textbf{Tiny-}\\ \textbf{ImageNet}} & \textbf{PACS} & \makecell{\textbf{Office-}\\ \textbf{Caltech-10}} \\
    
    \hline
    MA-Echo & 10.03$\pm$1.38 & 0.51$\pm$0.02 & 16.70$\pm$0.60 & 11.67$\pm$2.39 \\
    DFedAvgM & 18.59$\pm$1.65 & 2.02$\pm$0.20 & 21.58$\pm$2.28 & 10.01$\pm$0.70 \\
    DFedSAM & 18.51$\pm$1.28 & 3.15$\pm$0.29 & 20.79$\pm$1.36 & \boldmath{}\textbf{15.09$\pm$1.04}\unboldmath{} \\
    \hline
    \makecell{\method\\(PFL)} & \boldmath{}\textbf{24.71$\pm$2.62}\unboldmath{} & \boldmath{}\textbf{4.12$\pm$0.94}\unboldmath{} & \boldmath{}\textbf{24.13$\pm$1.57}\unboldmath{} & 12.66$\pm$1.36 \\
    \myhline
    \end{tabular}%
    }
  \label{tab:mesh}%
\end{table}%

      
    

\vspace{3mm}
\noindent\textit{D.7 PFL Adaptation.} Although our method is not exclusively designed for PFL, we have nevertheless adjusted our algorithm to fit the decentralized PFL setup for a fair comparison. As shown in Table \ref{tab:mesh}, it is evident that our method, even when adapted to the decentralized PFL structure, secures optimal results on most of the datasets relative to other baselines (we compare with \textbf{MA-Echo}~\cite{su2023one}, which represents the state-of-the-art one-shot decentralized PFL method; DENSE and FedOV which do not conform to the decentralized structure, are excluded from the comparison). It also surpasses most existing methods on the Office-Caltech-10 dataset. Therefore, our method successfully adapts to various settings and data distributions, highlighting its broad applicability.


\end{document}